\documentclass{article} 
\usepackage{conference,times}

\usepackage{amsmath,amsfonts,bm}

\def\eqref#1{equation~\ref{#1}}

\def\1{\bm{1}}

\DeclareMathAlphabet{\mathsfit}{\encodingdefault}{\sfdefault}{m}{sl}
\SetMathAlphabet{\mathsfit}{bold}{\encodingdefault}{\sfdefault}{bx}{n}

\DeclareMathOperator*{\argmin}{arg\,min}

\usepackage{hyperref}
\usepackage{url}
\usepackage{graphicx}
\usepackage{wrapfig}
\usepackage{booktabs}
\usepackage{multirow}
\usepackage{adjustbox}
\usepackage{array}
\usepackage{algorithm} 
\usepackage{algorithmic}
\newcolumntype{C}{>{\centering\arraybackslash}p{2cm}}

\usepackage{amsmath}
\usepackage{amssymb}
\usepackage{amsthm}
\usepackage{caption}
\usepackage{lipsum}
\newcommand{\eg}{\textit{e.g.}}
\newcommand{\ie}{\textit{i.e.}}
\newtheorem{proposition}{Proposition}[] 

\title{Open-Set Graph Anomaly Detection via \\ Normal Structure Regularisation}

\author{%
  Qizhou Wang\textsuperscript{1}, Guansong Pang\textsuperscript{2}\thanks{Corresponding author: Guansong Pang (gspang@smu.edu.sg)}, Mahsa Salehi\textsuperscript{1},  Xiaokun Xia\textsuperscript{3}, Christopher Leckie\textsuperscript{4} \\
  \textsuperscript{1}Monash University\\
  \textsuperscript{2}Singapore Management University\\
  \textsuperscript{3}The University of Tokyo\\
  \textsuperscript{4}The University of Melbourne\\
  \texttt{\{qizhou.wang, mahsa.salehi\}@monash.edu,  gspang@smu.edu.sg}  \\
  \texttt{xia.xiaokun@ipmu.jp,  caleckie@unimelb.edu.au}  \\
}

\finalcopy 
\begin{document}

\maketitle

\begin{abstract}
This paper considers an important Graph Anomaly Detection (GAD) task, namely open-set GAD, which aims to train a detection model using a small number of normal and anomaly nodes (referred to as \textit{seen anomalies}) to detect both seen anomalies and \textit{unseen anomalies} (\ie, anomalies that cannot be illustrated the training anomalies). Those labelled training data provide crucial prior knowledge about abnormalities for GAD models, enabling substantially reduced detection errors. However, current supervised GAD methods tend to over-emphasise fitting the seen anomalies, leading to many errors of
detecting the unseen anomalies as normal nodes.
Further, existing open-set AD models were introduced to handle Euclidean data, failing to effectively capture discriminative features from graph structure and node attributes for GAD.
In this work, we propose a novel 
open-set GAD approach, namely \textit{\underline{n}ormal \underline{s}tructure \underline{reg}ularisation} (\textbf{NSReg}),
to achieve generalised detection ability to unseen anomalies, while maintaining its effectiveness on detecting seen anomalies.
The key idea in NSReg is to introduce a regularisation term that enforces the learning of compact, semantically-rich representations of normal nodes based on their structural relations to other nodes. When being optimised with supervised anomaly detection losses, the regularisation term helps incorporate strong normality into the modelling, and thus, it effectively avoids overfitting the seen anomalies and learns a better normality decision boundary, largely reducing the false negatives of detecting unseen anomalies as normal. Extensive empirical results on seven real-world datasets show that NSReg significantly outperforms state-of-the-art competing methods by at least 14\% AUC-ROC on the unseen anomaly classes and by 10\% AUC-ROC on all anomaly classes. 
\end{abstract}

\section{Introduction}
Detection of anomalous nodes in a graph is a crucial task in the context of Graph Anomaly Detection (GAD) \citep{akoglu2015graph, gad_survey}. Its popularity has been growing in recent years due to its wide range of real-world applications, such as detection of malicious users in social networks, illegal transactions in financial networks, and faults in sensor networks. There have been numerous GAD methods introduced \citep{akoglu2015graph, domi, cola,gad_survey,qiao2024deep}, with the majority of them designed as unsupervised approaches. However, they are often associated with high detection errors due to the lack of knowledge about anomalies.

\begin{wrapfigure}{r}{0.5\columnwidth}
\vspace{-5pt}
\centering
\includegraphics[width=2.8in]{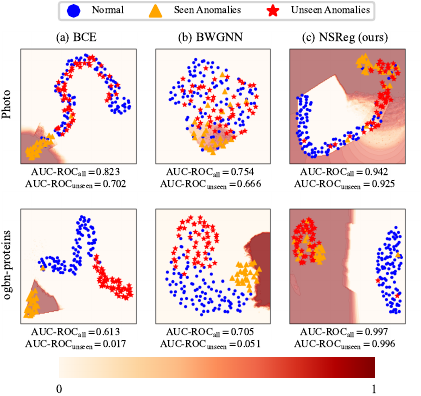}
\vspace{-10pt}
\caption{Visualisation of node representations with anomaly score contour lines of three supervised GAD models: baseline classifier Binary Cross-Entropy (BCE), recent state-of-the-art BWGNN \citep{bwgnn}, and our proposed NSReg, on two real-world datasets: Photo and ogbn-proteins. The models are trained using a limited number of seen anomalies and normal nodes. BCE and BWGNN overly focus on the decision boundary between the seen anomalies and normal nodes, whereas NSReg mitigates overfitting to the seen anomalies, resulting in more generalised detection of seen and unseen anomalies.}
\vspace{-10pt}
\label{fig:intro}
\end{wrapfigure}

There has been growing interest in supervised Anomaly Detection (AD) methods \citep{semi_ad_survey, pre, bwgnn,dev,pang2021toward} because they are able to utilise labelled anomaly data to substantially reduce the high detection errors of unsupervised approaches.
Such methods assume the availability of a limited number of labelled anomalies, which is usually feasible to obtain in real-world applications and can be utilised to enable anomaly-informed supervision. Despite their generally superior performance, these methods prove less effective in open-set AD where training models with labelled anomalies (referred to as \textit{seen anomalies}) fails to adequately represent anomalies at inference time, particularly in the context of newly emerging types/classes of anomalies that are substantially different from those seen in training (\ie, \textit{unseen anomalies}). This is because such methods often concentrate solely on modelling abnormal patterns derived from labelled anomalies, thus exhibiting \textit{poor generalisation to the unseen anomalies}, \ie, \textit{many unseen anomalies are detected as normal}, as shown in Figure \ref{fig:intro}(a-b). Further, they are mostly designed for handling Euclidean data like image and tabular data \citep{dev,pre}, thereby overlooking valuable discriminative information on the structure and node attributes in graph data for GAD.

To address these issues, this paper focuses on the practical yet under-explored problem of open-set GAD, aiming to enhance the generalisation for both seen and unseen anomaly nodes by leveraging a small
set of labelled seen anomaly nodes.
To this end, we propose a novel open-set GAD approach, namely \textit{\underline{n}ormal \underline{s}tructure \underline{reg}ularisation} (\textbf{NSReg}), to leverage the rich normal graph information embedded in the labelled nodes. The key idea in NSReg is to introduce a regularisation term that enforces the learning of compact, semantically-rich representations of normal nodes based on their structural relations to other nodes. When being optimised with supervised AD losses, the regularisation term incorporates strong normality into the modelling, and thus, it effectively avoids overfitting the seen anomalies, while learning better normality decision boundary. This helps substantially reduce the errors of detecting unseen anomalies as normal.

In particular, to capture those semantically-rich normal structure relations, NSReg differentiates the labelled normal nodes that are connected in their local neighbourhood from those that are not. This is done by predicting three types of normal-node-oriented relation prediction, including: {\it connected normal nodes}, {\it unconnected normal nodes}, and {\it unconnected normal nodes to unlabelled nodes}. We show theoretically that our regularisation module prioritises establishing distinct node representations based on their structural relationships with the normal class, rather than attempting to predict these anomalies without grounded information. As a result, it effectively reinforces the structural normality of the graph in the representation space and enforces a more stringent decision boundary for the normal class, enabling better separability of the unseen anomaly nodes from the normal nodes, as shown in Figure \ref{fig:intro}(c).
Moreover, NSReg is a plug-and-play module, which can be integrated as a plugin module into various supervised GAD learning approaches to enhance their generalisability to unseen anomalies. 
It is worth noting that NSReg is designed to enhance the generalisation of supervised GAD to unseen anomalies by leveraging discriminative structural information derived from a small number of labelled nodes. It is not intended for unsupervised GAD, where label information is unavailable, and models designed for the unsupervised setting are prone to different issues, \eg, the failure to capture anomaly patterns of interest or high false positives. 

In summary, our main contributions are as follows:
\begin{itemize}
    \item We study an under-explored problem of GAD, namely open-set GAD, and validate the feasibility of leveraging graph structural normality to regularise representation learning in supervised GAD, resulting in effective mitigation of the overfitting on the seen anomalies.
    \item We propose NSReg that regularises supervised GAD models by differentiating normal-node-oriented relations.
    It tackles the problem by enforcing better separation between unseen anomalies and normal nodes while retaining seen anomaly detection performance.
    \item NSReg is a plug-and-play regularisation term that can be applied to enhance the generalisation of different supervised GAD methods.
    \item We show through extensive empirical studies that NSReg significantly outperforms all competing methods by at least 14\% AUC-ROC on the unseen anomaly classes and by 10\% AUC-ROC on all anomaly classes.
\end{itemize}

\section{Related Work}
{\bf Graph Anomaly Detection.} GAD methods are usually designed as unsupervised learning to bypass the reliance of labelled data \citep{gad_survey, semi_ad_survey}. Earlier shallow methods \citep{anomalous, amen, coda, ryder, netwalk} leverage various heuristics to detect anomalies. Recent GAD methods predominantly utilise graph neural networks (GNNs) \citep{wu2020comprehensive} and have demonstrated superior performance due to their strong representation learning capacity. There are numerous unsupervised GNN-based GAD methods proposed \citep{domi, specae, meng_jiang_error, aegis, ocgnn, cola, conda, tam, he2024ada}.
However, these methods are not trained using real anomalies and have proven to be ineffective when their heuristic optimisation objective mismatches the actual anomaly patterns. To leverage anomaly-specific prior knowledge, supervised learning that utilises a small number of labelled anomalies \citep{abuse, bwgnn, amnet, caregnn, auc-oriented, DCI, ghrn} and learning schemes such as meta learning and transfer learning have also been explored for GAD \citep{meta-gdn, cmd, cd_gad, act}. Nevertheless, these methods are prone to overfitting the small number of labelled anomaly nodes and do not implement effective regularisation to ensure the generalisation on the unseen anomalies.
\smallskip

{\bf Towards Supervised Anomaly Detection.} Although numerous AD methods have been proposed \citep{chandola2009anomaly, ad_survey}, most of them \citep{zenati2018adversarially, park2020learning, roth2022towards} adopt optimisation objectives that focus on representation learning and are indirect for anomaly scoring. More recent works \citep{sohn2020learning, sad, dev, pre,pang2021toward,zhao2018xgbod} leverage a small number of labelled anomalies to perform optimisation and anomaly scoring in an end-to-end pipeline, significantly improving detection performance. However, such methods are originally designed for non-structured data and cannot explore structural information when directly applied for GAD.
\smallskip

{\bf Imbalanced Node Classification.} Another closely related line of research is imbalanced classification \citep{smote_ori, tkde_imbalance,  johnson2019survey}, which is a longstanding challenge in mitigating the class imbalance in the training data. A variety of approaches \citep{aug_survey, rect, qu2021imgagn, smote, ens} have been proposed to mitigate class imbalance in the training graph data to avoid overfitting the majority classes. However, these methods only assume a fixed number of known classes and do not consider the potential unseen classes at inference time. As a result, they fail to generalise to unseen anomaly classes in open-set GAD.

\begin{figure*}[t]
    \centering
    \includegraphics[width=1\linewidth]{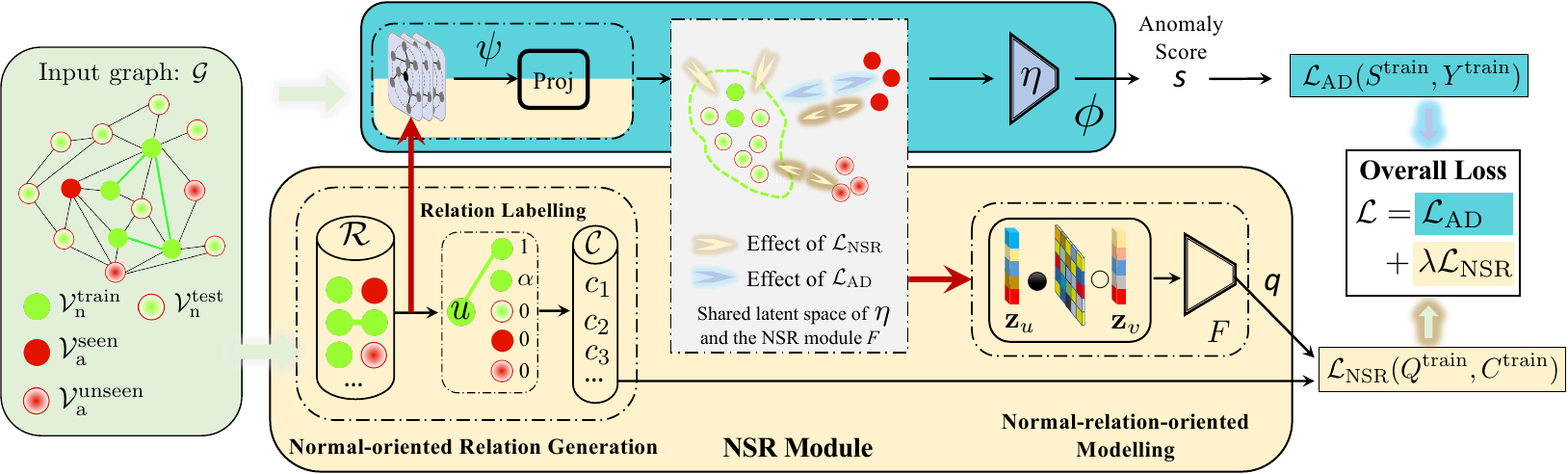}
    \caption{An overview of our proposed approach, NSReg, illustrating the integration
    of the NSR module
    into a graph anomaly detector as a plug-and-play module for regularising supervised GAD training, where red arrows indicate the ``plugging points". The green teal dashed box illustrates the decomposition of NSReg's overall learning objective in the shared representation space.
    Here, $\mathcal{L}_{\mathrm{NSR}}$ focuses on 
    enforcing a more stringent decision boundary for the normal class, enabling better separability of the unseen anomaly nodes from the normal nodes, which is not considered by $\mathcal{L}_{\mathrm{AD}}$.}
    \label{fig:overview}
    \vspace{-15pt}
\end{figure*}

\section{Our Proposed Approach}
\subsection{Problem Statement}
We consider open-set GAD on attributed graphs. Let $\mathcal{G} = (\mathcal{V}, \mathcal{E}, \mathbf{X})$ be an attributed graph with a node set $\mathcal{V}$, an edge set $\mathcal{E}$ and a feature matrix $\mathbf{X}$, which contains significantly fewer anomalous nodes $\mathcal{V}_{\mathrm{a}}$ than normal nodes $\mathcal{V}_{\mathrm{n}}$.
In open-set GAD, during training, the labelled training nodes $\mathcal{V}^{\mathrm{train}}$ are often unable to illustrate all possible anomaly classes at inference. For clarity, we use $\mathcal{V}_{\mathrm{a}}^{\mathrm{seen}}$ to denote anomaly nodes that can be illustrated by the labelled anomaly nodes (seen anomalies) and $\mathcal{V}_{\mathrm{a}}^{\mathrm{unseen}}$ to denote the unseen anomalies, such that  $\mathcal{V}_{\mathrm{a}} = \mathcal{V}_{\mathrm{a}}^{\mathrm{seen}} \cup \mathcal{V}_{\mathrm{a}}^{\mathrm{unseen}}$ and $\mathcal{V}^{\mathrm{train}}_{\mathrm{a}} \subset \mathcal{V}_{\mathrm{a}}^{\mathrm{seen}}$.

Our objective is to learn a scoring function $\phi: (\mathcal{G}, \mathcal{V}) \rightarrow \mathbb{R}$, such that $\phi(\mathcal{G}, v_{\mathrm{a}}) \gg \phi(\mathcal{G}, v_{\mathrm{n}})$, for all $v_{\mathrm{a}} \in \mathcal{V}_{\mathrm{a}}^{\mathrm{seen}} \cup \mathcal{V}_{\mathrm{a}}^{\mathrm{unseen}}$ and $v_{\mathrm{n}} \in \mathcal{V}_{\mathrm{n}}$. In this paper, we consider a GNN-based anomaly scoring function $\phi$, which consists of a pipeline with two components: a graph representation learner $\psi(\mathcal{G}, \mathcal{V}; \Theta_{\psi})$: $(\mathcal{G}, \mathcal{V}) \rightarrow \mathcal{Z}$ and an anomaly scoring function $\eta(\mathcal{Z}; \Theta_{\eta})$ : $\mathcal{Z} \rightarrow \mathbb{R}$, where $\Theta_{\psi}$ and $\Theta_{\eta}$ are learnable parameters. We aim to obtain the following anomaly scoring mapping:
\begin{equation}
    \phi(\mathcal{G}, \mathcal{V}; \Theta_{\phi}) = \eta(\psi(\mathcal{G}, \mathcal{V};\Theta_{\psi});\Theta_{\eta}),
\end{equation}
with the support of the labelled nodes and the graph structure. 
The main challenges are the unforeseeable nature of anomalies and the  risk of overfitting to labelled anomaly nodes.

\subsection{Overview of NSReg}
In this paper, we propose a novel open-set GAD framework, namely \underline{n}ormal \underline{s}tructure \underline{reg}ularisation (NSReg), as a solution to tackle the aforementioned challenges. The key insight behind NSReg is to explicitly guide the representation learner in capturing discriminative normal structural information from the graph characterised by the labelled normal nodes, supplementing the incomplete and often biased label information provided by the labelled anomalies that represent the seen anomalies. The modelling of this structural normality effectively calibrates the normality decision boundary, enabling better generalisation to the unseen anomalies, \ie, better separation between the unseen anomalies and normal nodes. 

As illustrated in Figure \ref{fig:overview}, NSReg leverages a novel Normal Structure Regularisation (NSR) module to enforce compact and uncluttered normal subspace in the representation space. During training, the NSR module is integrated as a regularising component with a supervised graph anomaly detector, both sharing the same representation learner for joint optimisation, to enable the joint learning of the structural normality and seen anomaly patterns for GAD. The general objective of NSReg can be defined as the following:
\begin{equation}
\argmin_{\Theta_{\phi}, \Theta_{\mathrm{NSR}}}\sum_{v \in \mathcal{V}} \mathcal{L}_{\mathrm{AD}}(s_v, y_v) + \lambda \sum_{r\in \mathcal{R}} \mathcal{L}_{\mathrm{NSR}}(q_{r}, c_{r}),
\end{equation}
 where $\Theta_{\phi}$ and $\Theta_{\mathrm{NSR}}$ are the respective learnable parameters of the anomaly detection $\phi$ and the NSR module, and $\mathcal{L}_{\mathrm{AD}}$ is the loss function of a supervised GAD model, with $s_v$ being the predicted anomaly score and $y_v$ being the ground truth. Additionally, $\mathcal{L}_\mathrm{NSR}$ signifies the loss function of the NSR module adjusted by the regularisation coefficient $\lambda$, with $q_{r}$ denoting the predicted normality of a relation sample $r$, and its normality $c_r$ is defined by a labelling function $C$.
 During inference, the representation learner $\psi*$ is combined with the trained anomaly scoring network $\eta*$ to form an end-to-end pipeline for detection, and the NSR module is disconnected from the representation learner since the normal structural information has already been learned. This simplifies the inference process without introducing any additional NSR module-related runtime or resource overhead.

\subsection{Normal Structure Regularisation (NSR)}
\label{sec:nsr}
The design of the NSR module, which is the core of NSReg, is motivated by the limitations of most supervised GAD methods, which are only designed to maximise separability between normal nodes and seen anomalies, but fail to provide sufficient supervision for the representation learner to effectively differentiate unseen anomaly representations from the normal class.
In an open-set environment, we are unable to obtain the prior knowledge of the unseen anomalies, and thus, difficult to learn the unseen anomaly patterns. Thus, NSReg takes a step back and focuses on learning better normality, which would help distinguish the unseen anomalies from the normal nodes better. NSReg achieves this by modelling the normal-node-oriented relations  (i.e., $\{r=(v, u) \mid v \in \mathcal{V}_{\mathrm{n}}, u \in \mathcal{V}\}$), which is aimed at enforcing a stricter definition of the normal region and recalibrating misplaced unseen anomaly representations within the representation space. By modelling three types of normal-node-oriented relations as a discriminative task, NSReg enhances representation learning with significantly enriched normality semantics, effectively disentangling unseen anomaly nodes from normal nodes in the representation space. 
We first provide a theoretical analysis of enforcing structural normality and then detail the two core components of NSR: normal-node-oriented relation generation and modelling.
\medskip \\ 
\noindent{\bf Preserving Structural Normality Improves Representation Learning for GAD.} We analyse the effects of enforcing structural normality in the representation space and its advantages for enhancing generalisation to unseen anomalies. We consider a mapping \( g:(\mathbb{R}^d, \mathbb{R}^d) \rightarrow \mathbb{R}_{[0,1]} \) in the node representation space \(\mathcal{Z}\), which takes the representations of nodes in each relation as inputs to model their normality. The normality of the relations is defined by a labelling function \( C \), where 0 and 1 indicate the lowest and the highest levels of normality, respectively. 
$C$ should satisfy that $0 <= \max(\mathcal{C}_{\mathrm{a}}) \ll  \min(\mathcal{C}_{\mathrm{n}}) <=1$,
where $\mathcal{C}_{\mathrm{n}}$ and $\mathcal{C}_{\mathrm{a}}$ indicate the respective set of the labels for the relations between only the normal nodes and the relations between normal and anomaly nodes respectively, emphasising a significant differentiation in the scale of normality. 

Specifically, \( g = g_C \circ g_E \) consists of two sub-mappings: \( g_E \), a linear mapping that fuses the node representations to produce relation representations, with the option to incorporate a scaling function on these node representations; and \( g_C \), a linear mapping with a monotonic final activation function to assign normality scores based on the representation. Note that both the scaling function and activation function are required to be homeomorphic.
If $\mathcal{Z}$ is shared with a discriminative graph anomaly detector, the normal nodes and the observed anomalies will be well separated into two isolated dense regions. Based on this observation, we demonstrate through the following proposition that differentiating between these two types of relations will force any anomalous nodes, including unseen ones, to be excluded from the normal region.
\begin{proposition}\label{prop1}
Consider a well-trained mapping $g$ that effectively distinguishes between the two types of normal-node-oriented relations within a relation representation space \( \mathcal{H} \), derived from a node representation space \( \mathcal{Z} \). The first type consists exclusively of relations among normal nodes, while the second type involves one normal and one anomalous node, with normality defined by some labeling function \( C \), regardless their connectivity. Consider the subspace of all normal nodes in $\mathcal{Z}$ as a connected open set $\mathcal{Z}_{\mathrm{n}}$, the boundary of which is defined by some closed hypersurfaces $\mathcal{M} = \{M_i \vert i \in \{1,\cdots,k\}\}$. Furthermore, let $\mathcal{Z}_{\mathrm{m}}$ denote the union of the interior of each $M_i \subset \mathcal{M}$. Given such $\mathcal{Z}_{\mathrm{n}}$ and $\mathcal{Z}_{\mathrm{m}}$, we can obtain that $\mathbf{z} \in \mathcal{Z}_{\mathrm{n}}$ is true for all $\mathbf{z} \in \mathcal{Z}_{\mathrm{m}}$, indicating their equivalence.
\end{proposition}
The proof, along with an intuitive diagram (Figure \ref{fig:apd:p1}), is provided in Appendix \ref{sec_apd_p1}. There are two key insights we can obtain from this proposition. First, since it is obvious that $\mathcal{Z}_{\mathrm{n}}$ is a subset of $\mathcal{Z}_{\mathrm{m}}$, by proving their equivalence, we observe that inside the boundary of the normal subspace, no anomalous subspace exists, and therefore, no anomaly nodes, whether seen or unseen, are present.
Second, this implies that for normal-node-oriented relations involving anomalies to be distinguishable from relations exclusively between normal nodes, the only way to adjust the representation learner is to ensure that no anomaly node resides within the normal space \(\mathcal{Z}_{\mathrm{n}}\). Therefore, enforcing structural normality will result in the displacement of misplaced anomalies from the normal subspace. These insights motivate the design of our NSR module, which is detailed below.
    
\noindent{\bf Normal-node-oriented Relation Generation.} The relation generation module implements the labelling function \( C \) and oversees relation sampling, specifically tailored to integrate structural normality knowledge in open-set GAD. It first samples normal-node-oriented relations and defines their normality, considering three types of relations, including connected normal relations $\mathcal{R}_{(\mathrm{n},\mathrm{c}, \mathrm{n})}$, 
unconnected normal relations $\mathcal{R}_{(\mathrm{n},\mathrm{u}, \mathrm{n})}$, and unconnected normal to other nodes, $\mathcal{R}_{(\mathrm{n},\mathrm{u}, \mathrm{u})}$.  A labelling function 
$C$, which meets the requirements outlined in Proposition 1, is then used to define their normality scores as follows:
\[
C(r) = 
\begin{cases}
   \text{1} & \text{if } r \in \mathcal{R}_{(\mathrm{n},\mathrm{c}, \mathrm{n})} =  \{(v,u) \mid v, u \in \mathcal{V}^{\mathrm{train}}_{\mathrm{n}}, (v, u) \in \mathcal{E}\}  \\
   \alpha & \text{if } r \in \mathcal{R}_{(\mathrm{n},\mathrm{u}, \mathrm{n})} =  \{(v,u) \mid v, u \in \mathcal{V}^{\mathrm{train}}_{\mathrm{n}}, (v, u) \notin \mathcal{E}\}   \\
   \text{0} & \text{if }r \in \mathcal{R}_{(\mathrm{n},\mathrm{u}, \mathrm{u})}  = \{(v,u) \mid v \in \mathcal{V}^{\mathrm{train}}_{\mathrm{n}}, u \in \mathcal{V} \setminus \mathcal{V}^{\mathrm{train}}, (v, u) \notin \mathcal{E}\},\\
\end{cases}
\]
where $ 0 \ll \alpha < 1$ is the specification of the normality of $\mathcal{R}_{(\mathrm{n},\mathrm{u}, \mathrm{n})}$ relative to other two types of relations. $\alpha=0.8$ is set by default to define a three-level normality hierarchy we want to enforce. This specification effectively embeds the homophilic assumption between the normal nodes while preserving the difference between the most related and related normal relations. We also find it unnecessary to include normal - seen anomaly pairs because similar information can be learned from supervised GAD objectives. Due to space limit, please refer to Appendix \ref{sec:apd:na} for details.
\medskip \\
\noindent{\bf Normal-node-oriented Relation Modelling.} This module instantiates the discriminative mapping \( g \) for normal relation modeling as a neural network, denoted \( F \). It first generates the representations of relations by fusing the representations of their corresponding end nodes using a learnable mapping \( F_E \), which is then fed into a normality prediction network \( F_C \) to model their normality. Specifically, $F_E$ can be defined as: 
\begin{equation}
\mathbf{h}_{r} = F_E \big( \psi(\mathcal{G},v), \psi(\mathcal{G}, u) \big)= \sigma(\mathbf{z}_v) \cdot \mathbf{W}_{E}  \circ \sigma(\mathbf{z}_u),
\end{equation}
where $\sigma$ is the sigmoid function, and $\mathbf{W}_E$ is a learnable weight matrix. $\mathbf{z}_v$ and $\mathbf{z}_u$ are the representations of the node in relation $r$, generated by the node representation learner $\psi$, which comprises a GNN-based representation learner followed by a projection network. The default GNN is a two-layer GraphSAGE \citep{sage} considering its good learning capacity and scalability. 
The relation modelling network is then optimised according to the following loss function:
\begin{equation}
        \mathcal{L}_{\mathbf{NSR}} = - \Big(c_{r} \cdot \log\big(F_C(\mathbf{h}_r)\big) + (1 - c_{r}) \cdot \log\big(1 - F_C(\mathbf{h}_r)\big)\Big),
\end{equation}
where $c_r$ is the relation label subject to the labelling function $C$ and $F_C$ is implemented as a learner layer followed by the Sigmoid function. This design satisfied all the conditions required for Proposition \ref{prop1}, ensuring that the normality enforced by the NSR module is accurately reflected in the shared representation space. It produces distinctive node representations in relation to those of the labelled normal nodes, thereby enforcing significantly enriched, fine-grained normality among the node representations. This enhancement directly regularises the representation learning of supervised anomaly detectors, enabling better separation of the unseen anomaly nodes from the normal nodes.

Note that our NSR module is different from PReNet \citep{pre}. PReNet is a weakly-supervised anomaly detector trained with an anomaly-oriented relation network, whereas NSR is a regularisation term focusing on learning fine-grained normal representations to regularise supervised anomaly detectors.

\subsection{Open-set GAD using NSReg}
\noindent{\bf Training.}\label{sec:training}
NSReg is trained through batch gradient descent over a predefined number of iterations. At each iteration, we first sample a batch of $b_{\mathrm{AD}}$ training nodes $V$, which contains $b^{\mathrm{n}}_{\mathrm{AD}}$ labelled normal nodes and all labelled anomaly nodes (note that the number of labelled anomalies is typically very small) for tuning the anomaly scoring network $\eta$ and the representation learner $\psi$. In addition, the normal-node-oriented relation generation is performed to generate $b_{\mathrm{NSR}}$ relation samples $R$ for optimising the NSR module and $\psi$. The overall training objective of NSReg can be formulated as:
\begin{equation} \label{eqn:overall}
    \argmin_{\Theta_{\phi}, \Theta_{\mathrm{NSR}} }
    \frac{-1}{{\vert V \vert} }\big(Y_{V} \log(S_{V}) + (1-Y_{V}) \log\big(1-S_{V})\big) + \frac{\lambda}{{\vert R \vert}} \mathcal{L}_{\mathrm{NSR}}
    (Q_{R}, C_{R}), 
\end{equation}
where the first term is a simple supervised cross-entropy-loss-based anomaly detector and the second term is our NSR module. As shown in Sec. \ref{sec:main_result}, the NSR term can be also effectively combined with other supervised GAD models in the first term.
During training, the GAD loss is computed initially based on the first term of the objective function and is used to update the parameters of the anomaly scoring network $\eta$. Subsequently, the NSR loss can be calculated using the second term of the objective function to update the relation modelling network. Finally, the combined GAD and NSR losses are aggregated and backpropagated to update the parameters of the representation learner. A Python-style pseudocode for training is provided in the Appendix \ref{pesudo}.

\noindent{\bf Inference.} During inference, given a test node $v$, NSReg first generates its representation $\mathbf{z}_v$, which is then scored using the trained anomaly scoring network $\eta$: $ s_v = \phi(\mathcal{G}, v) = \eta(\psi(\mathcal{G}, v))$.

\section{Experiments}\label{sec:exps}
{\bf Datasets.}
NSReg is extensively evaluated using seven real-world attributed graph datasets. To the best of our knowledge, no publicly available GAD datasets include multiple types of human-annotated natural anomaly classes. Therefore, we adapt imbalanced node classification and binary-labelled GAD datasets to define anomaly subclasses with distribution discrepancies for open-set GAD evaluation. For imbalanced node classification datasets with multiple minor classes, such as {\it Photo}, {\it Computers}, and {\it CS} \citep{pitfalls}, we treat each minor class (those with less than 5\% of total nodes) as seen anomalies, and the rest as unseen anomalies. This is consistent with the general definition of anomalies, characterised by rarity and significant difference from the majority. In the case of {\it Yelp} \citep{yelp_leman} and {\it T-Finance} \citep{bwgnn}, two well-known GAD datasets with binary labels, we cluster the learned representations of the anomaly class to create anomaly subclasses. Three large-scale attributed graph datasets, {\it ogbn-arxiv}, {\it ogbn-proteins} \citep{ogb}, and T-Finance \citep{bwgnn} are also adapted to evaluate NSReg at scale. More details about the datasets are presented in Appendix~\ref{sec:abl:dset}. 

This approach provides a realistic and comprehensive evaluation of detecting real-world anomalies, while also introducing an open-set GAD scenario with guaranteed semantic deviation across multiple anomaly classes. 
There are GAD datasets with injected anomalies using added artificial edges or node attributes in some earlier GAD works, but, following previous studies \citep{tang2023gadbench,qiao2024generative}, they are not used here since their injection patterns are often unintentionally incorporated into the design of detection models, resulting in unwanted information leakage. 

{\bf Evaluation Protocol and Metrics.}
For each dataset, we treat one of the anomaly classes as the seen anomaly, with the other anomaly classes as unseen anomalies.
We alternate this process for all anomaly classes, and report the results averaged over all cases. All experiments are repeated for 5 random runs. For example, the CS dataset includes 8 anomaly classes, each is treated as `seen' in turn, resulting in 8 sub-experiments per run, with a total of 40 sub-experiments across 5 runs. The Python style code for the evaluation protocol is presented in Appendix \ref{apd:eval}. 

We employ two widely used and complementary performance metrics in GAD: the Area Under Receiver Operating Characteristic Curve (AUC-ROC) and the Area Under Precision-Recall Curve (AUC-PR). 
In particular, AUC-ROC measures both true positives and false positives, while AUC-PR summarises the precision and recall of the anomaly classes, offers a focused measure exclusively on the anomaly classes.

{\bf Competing Methods.}  NSReg is compared with 16 competing methods from multiple related areas, including four popular unsupervised methods, eight state-of-the-art (SOTA) supervised methods and one baseline method. Specifically, DOMINANT \citep{domi}, GGAN \citep{ggan}, \citep{ocgnn}, CoLA \citep{cola}, \citep{conda}, TAM \citep{tam} and ADA-GAD \citep{he2024ada} are implemented as our unsupervised baselines due to their popularity and competitive performance among unsupervised GAD methods. The SOTA methods consist of recently proposed supervised anomaly detectors for Euclidean data: DevNet \citep{dev}, PReNet \citep{pre}, and those for graph data: DCI \citep{DCI}, BWGNN \citep{bwgnn}, AMNet \citep{amnet} and GHRN \citep{ghrn}, and two imbalanced node classification methods as well: GraphSMOTE \citep{smote} and GraphENS \citep{ens}.
The classification method using binary cross-entropy (BCE) is considered as a baseline model.

{\bf Implementation Details.}
NSReg comprises a representation learner featuring a two-layer GraphSAGE \citep{sage}, followed by a two-layer projection network, each containing 64 hidden units per layer. NSReg is optimised using the Adam \citep{adam} optimiser for 200 epochs for the Photo, Computers, and CS datasets with a learning rate of $1 \times 10^{-3}$, and for 400 epochs for Yelp with a learning rate of $5 \times 10^{-3}$ due to its larger number of nodes. We set the number of labelled anomalies to 50 and the percentage of labelled normal nodes to 5\% by default. Each batch of training nodes includes all labelled anomalies and randomly sampled number of labelled normal nodes capped at 512. Similarly, each batch of relations is capped at 512, with an equal number of samples for each relation type. The default value of $\alpha$ is set to 0.8 and $\lambda$ is set to 1 for all datasets. The analysis of their sensitivity can be found in Appendix \ref{app:hyperparameter}, where NSReg maintains stable performance, showing its robustness to the choice of hyperparameters. Further details are provided in Appendix \ref{apd:implem}. In our default setting, 50 anomalies are used for training along with 5\% of randomly selected normal nodes, while the remaining data is reserved for evaluation.

\begin{table*}[t]
\centering
\caption{AUC-ROC and AUC-PR (mean±std) for detecting all anomalies and exclusively unseen anomalies. The boldfaced are the best results. ``-'' denotes unavailable result due to out-of-memory.}
\vspace{-10pt}
\begin{adjustbox}{width=\linewidth}
\begin{tabular}{ccccccCccccC}
\toprule
                        &            & \multicolumn{5}{c}{AUC-ROC}                                         & \multicolumn{5}{c}{AUC-PR}                                         \\ \cmidrule(lr){3-6} \cmidrule(lr){7-7} \cmidrule(lr){8-11}  \cmidrule(lr){12-12} 
Test Set                & Method     & Photo       & Computers   & CS          & Yelp        & Average     & Photo       & Computers   & CS          & Yelp        & Average     \\ \midrule
\multirow{17}{*}{\begin{tabular}[c]{@{}c@{}}All-\\ Anom.s\end{tabular}}   
&DOMINANT &0.429±0.001  &0.576±0.007    &0.402±0.000     &0.609±0.003   &0.504±0.003    &0.072±0.000	&0.184±0.005  &0.187±0.000	&0.221±0.003  &0.166±0.002 \\
&GGAN     &0.435±0.003	&0.566±0.008    &0.396±0.005     &0.323±0.007   &0.430±0.006    &0.078±0.001	&0.177±0.005  &0.186±0.001	&0.103±0.001  &0.136±0.002 \\
&OCGNN  &0.523±0.049	&0.506±0.013	&0.509±0.015	&0.445±0.083	&0.496±0.034 &0.095±0.012	&0.158±0.008	&0.238±0.006	&0.029±0.008	&0.130±0.003 \\
&COLA     &0.825±0.033	&0.630±0.020    &0.481±0.015     &0.302±0.013   &0.560±0.020    &0.246±0.034	&0.233±0.024  &0.253±0.011	&0.103±0.003  &0.209±0.018 \\
&CONDA  &0.529±0.000	&0.523±0.000	&0.403±0.041	&0.309±0.017   &0.441±0.018    &0.106±0.000	&0.150±0.000	&0.193±0.011	&0.107±0.003	&0.139±0.005\\ 
&TAM      &0.626±0.004  &0.435±0.002	&0.637±0.009     &0.295±0.006	&0.498±0.005    &0.122±0.002	&0.131±0.001  &0.318±0.006	&0.138±0.087  &0.177±0.024 \\
&ADA-GAD  &0.391±0.010 &0.557±0.004 &0.310±0.040 &OOM &-  &0.069±0.003 &0.174±0.001 &0.163±0.009 &OOM &- \\
&PReNet   &0.698±0.019  &0.632±0.028	&0.547±0.016     &0.692±0.004	&0.642±0.017    &0.459±0.010	&0.374±0.031  &0.363±0.011	&0.336±0.006  &0.383±0.015  \\
&DevNet   &0.599±0.079   &0.606±0.064	 &0.769±0.029	&0.675±0.020 	&0.662±0.048	&0.223±0.155    &0.284±0.093  &0.684±0.018	&0.315±0.027  &0.375±0.073  \\
&DCI   &0.772±0.061	&0.683±0.051	&0.856±0.012	&0.689±0.059	&0.750±0.023 &0.452±0.099	&0.427±0.069	&0.635±0.028	&0.351±0.044	&0.466±0.031\\
&BWGNN    &0.728±0.026   &0.722±0.008	 &0.769±0.029	&0.727±0.012    &0.737±0.019    &0.313±0.066	&0.461±0.012  &0.687±0.048	&0.366±0.015  &0.457±0.035  \\
&AMNet    &0.773±0.001   &0.671±0.007	 &0.873±0.003	&0.695±0.011	&0.753±0.006    &0.487±0.003	&0.395±0.016  &0.784±0.004	&0.337±0.013  &0.501±0.009 \\
&GHRN &0.741±0.015	&0.604±0.023	&0.757±0.036	&0.713±0.021	&0.704±0.009 &0.360±0.034	&0.324±0.024	&0.615±0.056	&0.349±0.020	&0.412±0.016\\
&G. ENS   &0.712±0.005   &0.672±0.009	 &0.845±0.027	&0.572±0.011	&0.700±0.013    &0.246±0.008	&0.319±0.015  &0.515±0.016	&0.199±0.010  &0.320±0.012  \\
&G. SMOTE &0.616±0.043	 &0.700±0.046	 &0.731±0.009	&0.727±0.019	&0.694±0.029	&0.135±0.041	&0.369±0.043  &0.732±0.060	&0.300±0.019  &0.384±0.041 \\          
&BCE      &0.807±0.014	 &0.724±0.027	 &0.854±0.039	&0.712±0.017	&0.774±0.024	&0.515±0.011	&0.481±0.026  &0.756±0.006	&0.376±0.017  &0.532±0.015 \\
                        \cmidrule(lr){2-2} \cmidrule(lr){3-7} \cmidrule(lr){8-12}
& NSReg (Ours)       &\textbf{0.908±0.016}	&\textbf{0.797±0.015}	&\textbf{0.957±0.007}	&\textbf{0.734±0.012}	&\textbf{0.849±0.013}	&\textbf{0.640±0.036}	&\textbf{0.559±0.018}	&\textbf{0.889±0.016}	&\textbf{0.398±0.014}	&\textbf{0.622±0.021}            \\ \midrule
\multirow{17}{*}{\begin{tabular}[c]{@{}c@{}}Unseen-\\ Anom.s\end{tabular}}   
&DOMINANT &0.428±0.002	&0.576±0.008	&0.401±0.000	&0.633±0.004	&0.510±0.004  &0.041±0.000	&0.156±0.006	&0.169±0.000	&0.135±0.002	&0.125±0.002 \\
&GGAN     &0.435±0.006	&0.566±0.009	&0.395±0.005	&0.319±0.008	&0.429±0.007  &0.046±0.001	&0.150±0.004	&0.168±0.001	&0.055±0.000	&0.105±0.002 \\
&OCGNN  &0.523±0.050	&0.506±0.013	&0.509±0.015	&0.445±0.100	&0.496±0.040 &0.054±0.008	&0.133±0.010	&0.217±0.006	&0.016±0.005	&0.105±0.002\\
&COLA     &0.826±0.034	&0.629±0.024	&0.482±0.015	&0.291±0.015	&0.557±0.022  &0.156±0.029	&0.201±0.015	&0.232±0.011	&0.055±0.002	&0.161±0.014\\
&CONDA &0.525±0.000	&0.523±0.000	&0.402±0.041	&0.296±0.017	&0.437±0.018			&0.061±0.000	&0.127±0.000	&0.174±0.010	&0.058±0.003	&0.105±0.005\\
&TAM      &0.621±0.004	&0.435±0.002	&0.638±0.009	&0.293±0.001	&0.497±0.004  &0.073±0.001	&0.110±0.001	&0.294±0.006	&0.053±0.000	&0.133±0.002\\
&ADA-GAD &0.392±0.015 &0.556±0.004 &0.308±0.040 &OOM &-  &0.040±0.002 &0.149±0.001 &0.147±0.008 &OOM &-\\
&PReNet   &0.460±0.042	&0.557±0.033	&0.497±0.016	&0.615±0.007	&0.532±0.025  &0.044±0.004	&0.205±0.032	&0.232±0.010	&0.129±0.005	&0.153±0.013            \\
&DevNet   &0.468±0.040	&0.537±0.083	&0.739±0.032	&0.621±0.026	&0.591±0.045  &0.045±0.005	&0.200±0.060	&0.606±0.021 	&0.142±0.022	&0.248±0.027             \\
&DCI  &0.614± 0.093	&0.629±0.061	&0.847±0.013	&0.637±0.059	&0.681±0.033	&0.083±0.038	&0.288±0.060	&0.558±0.035	&0.157±0.024	&0.272±0.015\\
&BWGNN      &0.598±0.008	&0.570±0.026	&0.829±0.030	&0.674±0.022	&0.668±0.022	&0.068±0.004	&0.286±0.014 	&0.620±0.060	&0.167±0.015	&0.285±0.023 \\
&AMNet    &0.603±0.004	&0.606±0.008	&0.860±0.004	&0.604±0.014	&0.668±0.008  &0.068±0.002	&0.237±0.010	&0.739±0.006	&0.143±0.009	&0.297±0.007 \\
&GHRN &0.611±0.014	&0.533±0.024	&0.729±0.038	&0.659±0.035	&0.633±0.011	&0.068±0.003	&0.181±0.020	&0.552±0.060	&0.148±0.019	&0.237±0.024\\
&G. ENS     &0.518±0.009	&0.610±0.009	&0.699±0.011	&0.536±0.017 	&0.591±0.012	&0.053±0.002	&0.232±0.012 	&0.447±0.017	&0.094±0.008	&0.207±0.010             \\
&G. SMOTE   &0.464±0.046	&0.643±0.056 	&0.835±0.043 	&\textbf{0.716±0.020}	&0.665±0.041	&0.044±0.004	&0.232±0.043 	&0.678±0.066 	&0.167±0.018	&0.280±0.033      \\ 
&BCE        &0.650±0.026	&0.662±0.015 	&0.866±0.005	&0.658±0.023 	&0.709±0.017	&0.070±0.007	&0.293±0.018 	&0.723±0.009	&0.170±0.015	&0.314±0.012 \\ \cmidrule(lr){2-2} \cmidrule(lr){3-7} \cmidrule(lr){8-12}
&NSReg (Ours)       &\textbf{0.836±0.031}	&\textbf{0.752±0.019} &\textbf{0.953±0.008}	&0.690±0.025 &\textbf{0.808±0.021}	&\textbf{0.221±0.057}	&\textbf{0.417±0.024}	&\textbf{0.866±0.021}	&\textbf{0.196±0.020}	&\textbf{0.425±0.031}             \\ 
\bottomrule
\end{tabular}
\end{adjustbox}
\label{tab:main}
\vspace{-10pt}
\end{table*}

\subsection{Main Results} \label{sec:main_result}
\label{sec: performance_all}
{\bf GAD Performance on Both Seen and Unseen Anomalies.} We first evaluate the performance of NSReg on detecting all test anomalies (\ie, $\mathcal{V} \setminus \mathcal{V}^{\mathrm{train}}$, including samples of both seen and unseen anomaly classes). The results are reported in Table \ref{tab:main} (the upper part). We can see that NSReg achieves consistently and significantly better performance than all competing methods across all datasets in terms of both AUC-ROC and AUC-PR. On average, NSReg demonstrates a significant performance advantage over all competing methods, among which the supervised baselines prove to be considerably more effective than the unsupervised ones. Specifically, for AUC-ROC, NSReg achieves an improvement of 9.7\% and 12.7\% compared to the best and second-best competing methods, BCE and AMNet, respectively.
Additionally, NSReg shows 16.9\% and 24.2\% improvement compared to these two methods in terms of AUC-PR. The significant overall performance gain is mainly attributed to NSReg's strong capability of detecting unseen anomalies, while at the same time achieving an effective fitting of the seen anomalies. 

\smallskip
\noindent{\bf GAD Generalisation to Unseen Anomalies.}
The capability of detecting unseen anomalies in the test data (i.e., $\mathcal{V}^{\mathrm{test}}_{\mathrm{n}} \cup \mathcal{V}^{\mathrm{unseen}}_{\mathrm{a}})$ is more indicative for evaluating a model's open-set GAD ability.
As shown in Table \ref{tab:main} (lower part), 
NSReg again achieves significantly better performance, compared to the competing methods. Notably, on average, NSReg exhibits a greater margin of improvement on the unseen anomaly classes, with a 14\% and 21\% improvement over the top-2 performing competing methods in terms of AUC-ROC. Similarly, a 35\% and 43\% performance gain over the top-2 performing competing methods is observed for AUC-PR. The substantially enhanced performance underscores NSReg's strong ability in reducing the errors of detecting unseen anomalies as normal.

\begin{wrapfigure}{r}{0.45\columnwidth}
    \centering
    \vspace{-10pt}
    \includegraphics[width=0.45\columnwidth]{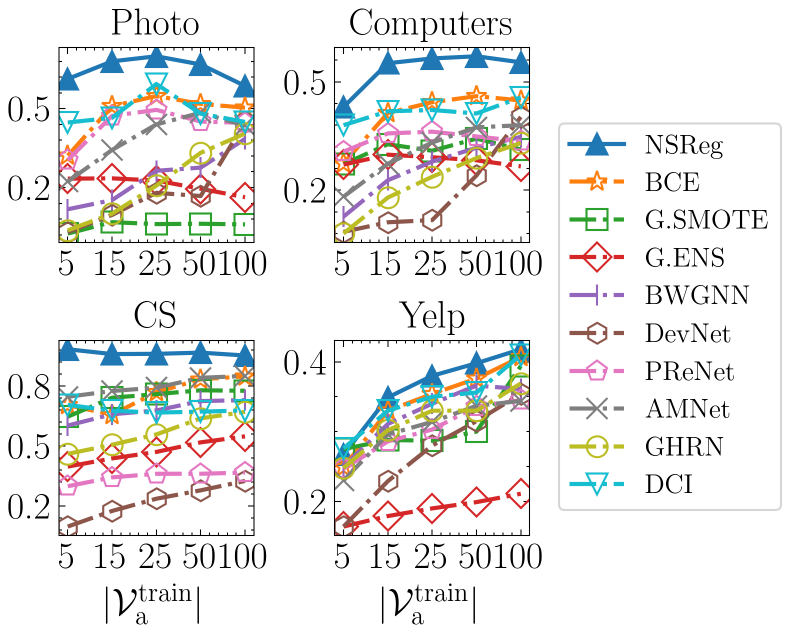}
    \vspace{-20pt}
    \caption{AUC-PR on all anomalies w.r.t. number of labelled anomalies.}
    \label{fig:de_aupr}
    \vspace{-10pt}
\end{wrapfigure}

\noindent\textbf{Data Efficiency.} We investigate NSReg's data efficiency by varying the number of seen anomalies around the default setting, including 5, 15, 25, 50, and 100, in order to understand the model's capacity to handle variations in data availability. The results are illustrated in Figure \ref{fig:de_aupr}.
Due to space constraints, we report results on the AUC-PR results on all test anomalies and provide the AUC-ROC results in Appendix \ref{apd:de}. 
The results show that NSReg consistently achieves significantly improved AUC-PR performance on the overall anomaly detection across the complete spectrum of the Photo, Computers, and CS datasets. On Yelp, NSReg either outperforms, or is on par with the competing methods across the entire range for overall performance. Similar findings can be observed on detecting unseen anomalies (see Figures \ref{fig:de_apd_aupr} and \ref{fig:de_apd_auroc}).

\begin{table}[t]
\begin{minipage}{0.45\textwidth}
\caption{Our NSR as a plugin module in supervised AD models DevNet, DCI and BWGNN.}
\begin{adjustbox}{width=1.0\columnwidth}
\begin{tabular}{cccccccc}
\toprule  
&Metric & \multicolumn{3}{c}{AUC-ROC} & \multicolumn{3}{c}{AUC-PR} \\  \cmidrule(lr){3-5} \cmidrule(lr){6-8} 
                           &Dataset           & Ori     & +NSR     & Diff.   & Ori     & +NSR     & Diff. \\ \midrule
\multirow{5}{*}{\rotatebox{90}{DevNet}} & Photo     & 0.599   & 0.684   & +0.085$\textcolor{blue}\uparrow$   & 0.223   & 0.424   & +0.201$\textcolor{blue}\uparrow$  \\
&Computers & 0.606   & 0.646   & +0.040$\textcolor{blue}\uparrow$   & 0.284   & 0.340   & +0.056$\textcolor{blue}\uparrow$  \\
& CS        & 0.769   & 0.949   & +0.180$\textcolor{blue}\uparrow$   & 0.684   & 0.872   & +0.188$\textcolor{blue}\uparrow$  \\
& Yelp      & 0.675   & 0.684   & +0.009$\textcolor{blue}\uparrow$   & 0.315   & 0.315   & +0.008$\textcolor{blue}\uparrow$  \\ \cmidrule(lr){2-2} 
 \cmidrule(lr){3-5} \cmidrule(lr){6-8} 
& Avg. Diff.  & \multicolumn{3}{r}{+0.079$\textcolor{blue}\uparrow$}   & \multicolumn{3}{r}{+0.113$\textcolor{blue}\uparrow$} \\ \cmidrule(lr){1-8}
\multirow{5}{*}{\rotatebox{90}{DCI}}  
&  Photo     &0.772	&0.865	&+0.093$\textcolor{blue}\uparrow$	&0.452	&0.577	&+0.125$\textcolor{blue}\uparrow$  \\
& Computers &0.683	&0.738	&+0.055$\textcolor{blue}\uparrow$	&0.427	&0.501	&+0.074$\textcolor{blue}\uparrow$  \\
& CS        &0.856	&0.857	&+0.001$\textcolor{blue}\uparrow$	&0.635	&0.623	&-0.012 \\
& Yelp      &0.689	&0.740	&+0.051$\textcolor{blue}\uparrow$	&0.351	&0.387	&+0.036$\textcolor{blue}\uparrow$ \\ \cmidrule(lr){2-2}  \cmidrule(lr){3-5} \cmidrule(lr){6-8} 
& Avg. Diff.& \multicolumn{3}{r}{+0.050$\textcolor{blue}\uparrow$}   & \multicolumn{3}{r}{+0.056$\textcolor{blue}\uparrow$}  \\ \cmidrule(lr){1-8}
\multirow{5}{*}{\rotatebox{90}{BWGNN}}  & Photo     &0.728	&0.802	&+0.074$\textcolor{blue}\uparrow$	&0.313	&0.530	&+0.217$\textcolor{blue}\uparrow$     \\
& Computers &0.635	&0.726	&+0.091$\textcolor{blue}\uparrow$	&0.348	&0.458	&+0.110$\textcolor{blue}\uparrow$  \\
& CS        &0.845	&0.879	&+0.034$\textcolor{blue}\uparrow$	&0.687	&0.718	&+0.031$\textcolor{blue}\uparrow$  \\
& Yelp      &0.727	&0.747	&+0.020$\textcolor{blue}\uparrow$	&0.366	&0.394	&+0.028$\textcolor{blue}\uparrow$  \\ \cmidrule(lr){2-2}  \cmidrule(lr){3-5} \cmidrule(lr){6-8}
& Avg. Diff.& \multicolumn{3}{r}{+0.055$\textcolor{blue}\uparrow$}   & \multicolumn{3}{r}{+0.097$\textcolor{blue}\uparrow$} \\ \bottomrule
\end{tabular}
\end{adjustbox}
\label{tab:improve}
\end{minipage}
\hfill
\begin{minipage}{0.5\textwidth}
\caption{Large-scale GAD results. ``-'' denotes unavailable result due to out-of-memory.}
\centering
\vspace{-2pt}
\begin{adjustbox}{width=1\textwidth}
\begin{tabular}{cccccc}
\toprule 
Metric &Datasets & ogbn-arxiv & ogbn-proteins &T-Finance & avg. \\ \midrule
\multirow{10}{*}{\begin{tabular}[c]{@{}c@{}c@{}}AUC-\\ ROC\\(All) \end{tabular}} 
&PReNet   &0.581±0.006	&0.613±0.035	&0.892±0.017	&0.695±0.019  \\
&DevNet   &0.601±0.015	&0.622±0.057	&0.654±0.210	&0.626±0.094  \\
&DCI      &0.566±0.041	&0.815±0.037	&0.763± 0.111	&0.715±0.063 \\
&BWGNN    &0.587±0.013	&0.727±0.067	&0.922±0.011	&0.745±0.030  \\
&AMNet    &0.600±0.049	&0.711±0.077	&0.889±0.024	&0.733±0.050  \\
&GHRN     &0.588±0.009	&0.674±0.019	&0.923±0.010	&0.728±0.013\\
&G.ENS    &-            &-          &0.847±0.087  &-          \\
&G.SMOTE  &-            &-          &0.875±0.016 &-          \\
&BCE      &0.592±0.004	&0.568±0.020	&0.922±0.011	&0.694±0.012    \\ \cmidrule(lr){2-6}  
&NSReg    &\bf{0.659±0.012}	&\bf{0.843±0.029}	&\bf{0.929±0.007} &\bf{0.810±0.016}  \\  \cmidrule{1-6}
\multirow{10}{*}{{\begin{tabular}[c]{@{}c@{}c@{}}AUC-\\ PR\\(All) \end{tabular}}} 
&PReNet  &0.288±0.004	&0.432±0.028	&0.571±0.140	&0.430±0.057   \\
&DevNet  &0.304±0.009	&0.436±0.017	&0.323±0.327	&0.354±0.118    \\
&DCI     &0.275±0.033	&0.597±0.062	&0.264±0.240	&0.379±0.112\\
&BWGNN   &0.263±0.009	&0.582±0.026	&0.746±0.023	&0.530±0.019\\
&AMNet   &0.272±0.053	&0.576±0.053	&0.644±0.046	&0.497±0.051\\
&GHRN    &0.271±0.007	&0.564±0.005	&0.727±0.031	&0.521±0.014\\
&G.ENS   &-           &-            &0.332±0.076    &-            \\
&G.SMOTE &-           &-            &0.573±0.077    &-            \\ 
&BCE     &0.310±0.003	&0.524±0.008	&0.726±0.034	&0.520±0.015   \\  \cmidrule(lr){2-6}
&NSReg   &\bf{0.336±0.006}	&\bf{0.723±0.020} &\bf{0.757±0.020} &\bf{0.605±0.125} \\
\bottomrule
\end{tabular}
\end{adjustbox}
\label{tab:large-scale}
\end{minipage}
\end{table}

\noindent\textbf{NSReg as a Plug-and-Play Module.}
This section examines the effect of using the NSR module of NSReg as a plugin module to enhance three other supervised methods, referred to as NSR-enabled models. 
Table \ref{tab:improve} shows the GAD performance on detecting all test anomalies, with the original BWGNN, DCI and DevNet as the baselines. Due to the page limit, the results for unseen anomalies are reported in Table \ref{tab:improve_apd_unseen}. We can observe that the NSR-enabled BWGNN, DCI and DevNet yield significant performance improvements across all datasets in both evaluation metrics. 
In particular, on average, it improves DevNet by over 10\% for AUC-ROC and 30\% for AUC-PR, for all anomalies and unseen anomalies. Similarly, for BWGNN, its NSR-variant outperforms the original model by over 7\% in terms of AUC-ROC on both test sets and more than 14\% for AUC-PR. For DCI, NSR also yields similar level of performance improvement. The consistent improvement demonstrates not only the high generalisability of our proposed NSR but also the importance of enforcing structural normality in supervised GAD. 

\noindent\textbf{Large-scale GAD Performance.}
\label{sec:large-scale}
We compare NSReg's GAD performance with supervised baselines on all test nodes and unseen anomalies of three large-scale GAD datasets, as presented in Table \ref{tab:large-scale} and Table \ref{tab:large-scale-apd-unseen} in Appendix \ref{sec:apd:large}. Notably, NSReg consistently demonstrates advantageous performance, as observed in other datasets in Table \ref{tab:main}. Moreover, NSReg exhibits superior efficiency in terms of data utilisation for large-scale GAD tasks—it achieves effective structural regularisation and significantly improves generalisation with only 100 seen anomalies, accounting for only less than 1\% of total anomalies.

\subsection{Ablation Study} \label{sec:abl}
This section empirically explores the effectiveness and significance of enforcing our structural normality within node representations by answering the following questions, with AUC-PR results reported in Table \ref{tab:abl} and a visualisations included in Figure \ref{fig:abl2}. We include the results in AUC-ROC in Table \ref{tab:abl_apd_auroc} in Appendix \ref{apd:ablation} due to space limits.

\noindent\textbf{How important is graph structural information when enforcing normality?} We answer this question by replacing the NSR module with the one-class hypersphere objective in Deep SAD (SAD) \citep{sad}, which minimises the volume of normal node representations to learn the normality without considering the graph structure information. Table 
 \ref{tab:abl} shows that NSR can significantly outperform the one-class learning SAD, highlighting the advantage of enforcing graph structure-informed normality. This is because, by simply tightening the normal representations without considering their structural relationships, SAD may not necessarily learn discriminative unseen anomaly representations, as illustrated in Figure \ref{fig:abl2}(SAD). We also found the one-class learning-based SAD is unstable due to its notorious risk of model collapse, while our NSR does not have this issue due to the fine-grained structural relation prediction.

\begin{table}[t]
\centering
\begin{minipage}[t]{0.43\linewidth}
    \captionof{figure}{Node representations learned via various regularisation schemes on Photo.}
    \includegraphics[width=\linewidth]{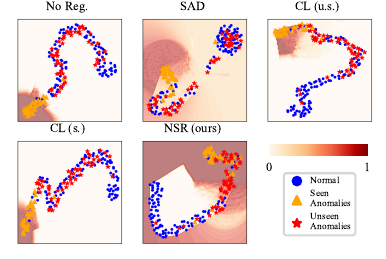}
    \label{fig:abl2}
\end{minipage}
\hfill
\begin{minipage}[t]{0.52\linewidth}
\caption{Our proposed NSR vs other regularisers in AUC-PR.}
\begin{adjustbox}{width=1\textwidth}
\begin{tabular}{cccccc}
\toprule
\multicolumn{6}{c}{All anomalies} \\ \cmidrule(lr){1-6}
Dataset    & SAD &CL (s.) &CL (u.s.) &NSR (dual)  & NSR (Ours) \\ \midrule 
Photo      &0.269±0.215 &0.513±0.018 &0.583±0.012 &0.623±0.029 &\bf{0.640±0.036} \\ Computers  &0.381±0.190 &0.492±0.014 &0.539±0.024	&0.539±0.023 &\bf{0.559±0.018} \\
CS         &0.700±0.016 &0.795±0.009&0.843±0.020	&0.858±0.042 &\bf{0.889±0.016} \\
Yelp       &0.175±0.046 &0.392±0.013	&0.330±0.078 &\bf{0.399±0.021} &0.398±0.014  \\ \cmidrule(lr){1-6}
average  	 &0.381±0.117 &0.548±0.014	&0.574±0.034 &0.605±0.029 &\bf{0.622±0.021}  
\\ \toprule
\multicolumn{6}{c}{Unseen anomalies} \\ \cmidrule(lr){1-6}
Dataset    & SAD &CL (s.) &CL (u.s.) &NSR (dual)  & NSR (Ours) \\ \midrule 
Photo     &0.053±0.004	&0.068±0.004	&0.133±0.019	&0.207±0.049	&\bf{0.221±0.057} \\
Computers &0.285±0.134	&0.318±0.018	&0.379±0.032	&0.389±0.028	&\bf{0.417±0.024} \\
CS        &0.631±0.019	&0.743±0.010	&0.797±0.025	&0.834±0.046 	&\bf{0.866±0.021} \\
Yelp      &0.090±0.017	&0.178±0.011	&0.152±0.045	&0.190±0.023 	&\bf{0.196±0.020}  \\ \cmidrule(lr){1-6}
Average   &0.265±0.044	&0.327±0.011	&0.365±0.030	&0.405±0.037	&\bf{0.425±0.031} 
\\ \bottomrule
\end{tabular}
\end{adjustbox}
\label{tab:abl}
\end{minipage}
\vspace{-30pt}
\end{table}

\noindent\textbf{Why is our proposed NSR more effective at structural regularisation?} To answer this question, we replace our NSR module with supervised and unsupervised contrastive learning (CL (s.) and CL (u.s.) in Table \ref{tab:abl}), which enforce similarities between nodes based on their connections. NSReg consistently outperforms both baselines on all tests for both metrics. This is because the NSR module is tailored to ensure that its enforced normality, which is normal-nodes-oriented and thus better aligned with the objective of enhanced normal class modelling, is reflected in the shared representation space, therefore resulting in a more stringent normal class subspace and significantly mitigating overfitting on the seen anomalies, as shown in Figure \ref{fig:abl2}(NSR).

\noindent\textbf{Can we just augment the seen anomalies to improve the generalisation on the unseen ones?} This can be answered by comparing the results of NSReg with graph augmentation-based methods G.SMOTE and G.ENS in Table \ref{tab:main}. NSReg significantly outperforms the two methods, suggesting that merely augmenting seen anomalies is inadequate for generalising to unseen ones. This is due to the fact that the augmented anomaly samples mainly resemble only the seen abnormalities, which may even further increase the risk of overfitting on the seen anomalies.

\smallskip
\noindent\textbf{Can we perform the relation prediction in NSR with less types of relations?} 
We compare NSR with its variant that is trained without the second relation $\mathcal{R}_{(\mathrm{n},\mathrm{u},\mathrm{n})}$ (NSR (dual) in short).
Table \ref{tab:abl} shows that the default NSR outperforms NSR (dual) by a non-trivial margin on almost all cases. This is because $\mathcal{R}_{(\mathrm{n},\mathrm{u},\mathrm{n})}$ is essential for defining our normality hierarchy, enabling finer granularity in relation modeling, preserving the structural differences between more closely related normal nodes and other nodes, on top of enforcing the distinction between normal and anomaly nodes.

\section{Conclusion and Future Work}
This paper introduces a novel open-set GAD approach, Normal Structure Regularisation (NSReg), which employs a plug-and-play regularisation term to enhance anomaly-discriminative normality from graph structures, thereby reinforcing strong normality in node representations during supervised GAD model training. By aligning node representations with their structural relationships relative to the normal nodes, NSReg establishes a stringent decision boundary for the normal class and enhances its generalisation to unseen anomalies by reducing the errors of detecting them as normal. Comprehensive empirical results show the superiority of NSReg in detecting both seen and unseen anomalies compared to 16 competing methods on several real-world graph datasets. 

In terms of limitations, despite its impressive effectiveness in improving generalisation in open-set GAD, NSReg models normality only within the immediate neighbourhood of the labelled normal nodes. While this is not an issue at present, as real-world interactions grow more complex and larger in scale, exploring normal relations beyond immediate neighbours to gain additional information for open-set GAD may become necessary. However, this could introduce additional computational overhead in the relation generation process. One potential solution is to implement efficient relation sampling or approximation strategies, which we leave for future work.

\bibliography{ref}
\bibliographystyle{conference}

\appendix
\section{Appendix}
\appendix
\bigskip
\textbf{\fontsize{15}{15}\selectfont Appendix} \smallskip \\
The appendix section is structured as follows. In Appendix \ref{sec:apd:proof}, we details the proof of the proposition 1. In Appendix  \ref{appendix:expdetails}, we provide extended details of the experimental setup, including dataset information, experimental protocol, and implementation specifics. In Appendix \ref{sec:apd:res} , we provide additional results that were not included in the main section due to space limitations. 

\section{Proof of Proposition 1}
\label{sec:apd:proof}
\begin{figure}[h]
    \centering
    \includegraphics[width=0.8\columnwidth]{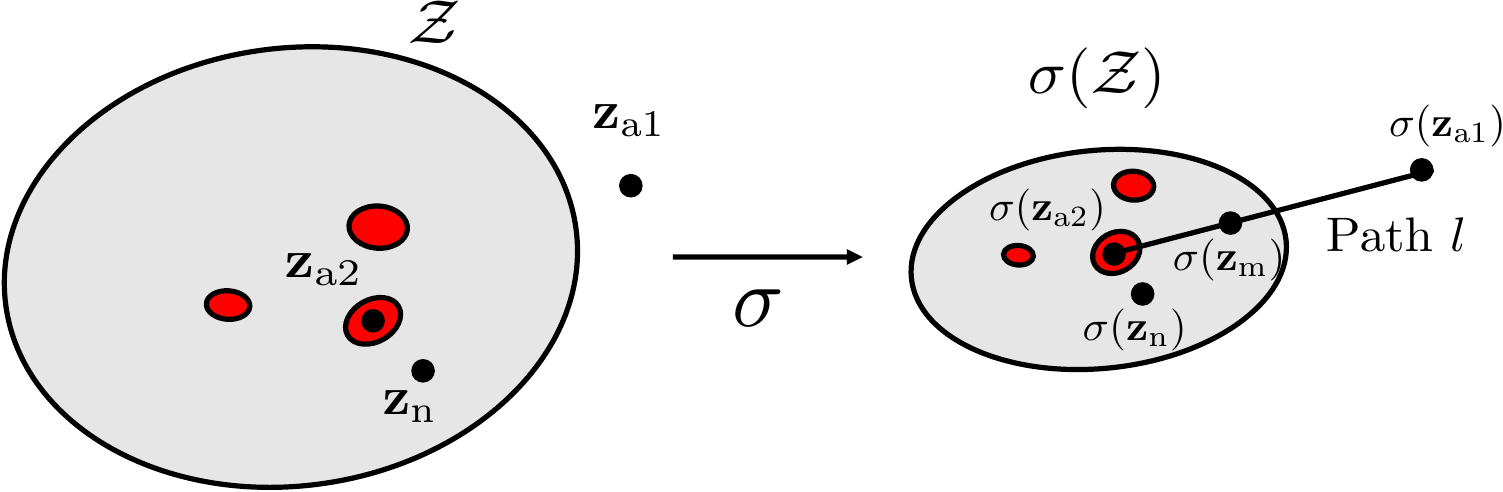}
    \caption{Proposition 1 proof intuition: $\mathcal{Z}$ and $\sigma(\mathcal{Z})$ represent the shared node representation spaces before and after applying the scaling function $\sigma$, respectively. In this plot, the grey region represents the normal subspace $\mathcal{Z}_{\mathrm{n}}$, and the union of the grey and red regions represents $\mathcal{Z}_{\mathrm{m}}$. The boundary of the normal subspace, $\mathcal{M}$, is depicted by the black lines. The goal is to prove that the red regions, which denote the neighbourhood of anomalies, do not exist within the normal subspace for a well-trained relation discriminator.}
    \label{fig:apd:p1}
\end{figure}

\label{sec_apd_p1}
For convenience, we define a labelling function $C$ with a range of $[0, 1]$. This function assigns a normality score of 0 to relations between normal nodes and anomaly nodes. In contrast, relations exclusively between normal nodes receive scores ranging from a significantly greater value than 0, denoted as $\alpha$, to a maximum of 1. For simplicity, we use the mapping of our default choice of relation modelling function $F$. It is worth noting that any function that satisfies the condition specified in Sec. \ref{sec:nsr} can be proved using the same approach. We show via proof by contradiction that, if $F$ is well-trained to distinguish the normal nodes exclusive relations and those that involves anomalies in node representation space shared with a supervised anomaly detector, no anomaly node lies in the normal region within the shared representation space of $\eta$ and $F$, where normal nodes form a dense subspace.

We assume the normal region in the shared representation space is a connected open set, due to the high similarity and dense distribution of normal representations forced by supervised discriminative loss $\mathcal{L}_{\mathrm{AD}}$. For a well-trained $F$ and any given normal node from the region $\mathrm{n}$, suppose there are two anomaly nodes, ${\mathrm{a}1}$ and ${\mathrm{a}2}$, where ${\mathrm{a}1}$ lies outside the normal region and ${\mathrm{a}2}$ overlaps with the normal region. Note that such $\mathrm{a}1$ can be easily found, as $\mathcal{L}_{\mathrm{AD}}$ will ensure a sufficient number of seen anomalies residing outside of the normal subspace $\mathcal{Z}_\mathrm{n}$. Consider a path connecting these two anomaly nodes:
\begin{equation}
    l_t = t\sigma(\mathbf{z}_{\mathrm{a}1}) + (1-t)\sigma(\mathbf{z}_{\mathrm{a}2}), t \in (0, 1).
\end{equation}
The homeomorphic of the Sigmoid function will ensure that we are able to find the scaled representation of a normal node $\sigma(\mathbf{z}_{\mathrm{m)}}$ along the path at $t = t_{\mathrm{m}}$. Then we have:
\begin{equation}
\sigma(\mathbf{z}_{\mathrm{m}}) = t_{\mathrm{m}}\sigma(\mathbf{z}_{\mathrm{a}1}) + (1-t_{\mathrm{m}} )\sigma(\mathbf{z}_{\mathrm{a}2}).
\end{equation}

Applying $F$ to the relation between $n$ and $m$, we have:

\begin{align*}
F(\mathbf{h}_{\mathrm{nm}}) &=\mathbf{W}_{C} \cdot \mathbf{h}_{\mathrm{nm}} + \mathbf{b} \\
&=\mathbf{W}_{C}(\operatorname{diag}(\sigma(\mathbf{z}_\mathrm{n})\cdot \mathbf{W}_{E})\sigma(\mathbf{z}_\mathrm{m}))+\mathbf{b}\\
&=\mathbf{W}_{C}(\operatorname{diag}(\sigma(\mathbf{z}_\mathrm{n})\cdot \mathbf{W}_{E})(t_\mathrm{m} \sigma(\mathbf{z}_\mathrm{a1})+(1 -t_\mathrm{m})\sigma(\mathbf{z}_\mathrm{a2}))+\mathbf{b}\\
&=t_\mathrm{m}\Big(\mathbf{W}_{C}\operatorname{diag}\big(
\sigma(\mathbf{z}_\mathrm{n})\cdot \mathbf{W}_{E}\big)\sigma(\mathbf{z}_\mathrm{a1})+\mathbf{b}\Big) + (1-t_\mathrm{m})\Big(\mathbf{W}_{C}\operatorname{diag}\big(\sigma(\mathbf{z}_\mathrm{n})\cdot \mathbf{W}_{E}\big)\sigma(\mathbf{z}_\mathrm{a2})+\mathbf{b}\Big) \\
&=t_\mathrm{m}\Big(\mathbf{W}_{C}\mathbf{h}_\mathrm{na1}+\mathbf{b}\Big) + (1-t_\mathrm{m})\Big(\mathbf{W}_{C} \mathbf{h}_\mathrm{na2}+\mathbf{b}\Big)
\\
&=t_\mathrm{m}F(\mathbf{h}_\mathrm{na1})+ (1-t_\mathrm{m})F(\mathbf{h}_\mathrm{na2}).
\end{align*}

Since $F$ is well trained, we have $F(\mathbf{h}_{\mathrm{na1}}) < \gamma-\delta$ and $F(\mathbf{h}_{\mathrm{na2}}) < \gamma-\delta$, where $\sigma(\gamma)=\alpha$ and $\delta \in \mathbb{R}_{(0, \gamma)}$. Therefore, we have :
\begin{align*}
    F(\mathbf{h}_{\mathrm{nm}}) &= t_\mathrm{m}F(\mathbf{h}_\mathrm{na1})+ (1-t_\mathrm{m})F(\mathbf{h}_\mathrm{na2}) < \gamma - \delta \\
     q_{\mathrm{nm}} &= \sigma(F(\mathbf{h}_{\mathrm{nm}}))< \sigma(\gamma - \delta)<\alpha
\end{align*}

which contradicts with the fact that $q_{\mathrm{nm}} > \alpha$. 

\section{More Experimental Details}\label{appendix:expdetails}
\subsection{Pseudocode for Training NSReg}
\label{pesudo}
The training procedure of NSReg is described in the following pseudocode:
\begin{algorithm}
\caption{Training NSReg}
\begin{algorithmic}[1]
\REQUIRE $\mathcal{G}$ and $Y^{\mathrm{train}} \text{ s.t. } Y^{\mathrm{train}} \ll \vert \mathcal{V} \vert $
\ENSURE $\phi: (\mathcal{G}, v) \rightarrow \mathbb{R}$ - an anomaly scoring mapping
\STATE Initialize variables
\FOR{$i=1$ in $[1,...,n_\mathrm{epochs}]$}
    \STATE $\mathbf{V} \leftarrow$ Sample $b_{\mathrm{AD}}$ nodes from $\mathcal{V}^{\mathrm{train}}$.
    \STATE $\mathbf{R}\leftarrow$ Sample $b_{\mathrm{NSR}}$ relations as described in Sec. \ref{sec:training}.
    \STATE $\mathcal{L}_{\mathrm{AD}} \leftarrow$ Compute the supervised GAD loss.
    \STATE Perform gradient descent for $\eta$.
    \STATE $\mathcal{L}_{\mathrm{NSR}} \leftarrow$ Compute the NSReg loss.
    \STATE Perform gradient descent for $F, E$.
    \STATE $\mathcal{L}\leftarrow$ Compute the total loss as in Eq.~\ref{eqn:overall}
    \STATE Perform gradient descent for $\psi$.
\ENDFOR
\RETURN $\phi$
\end{algorithmic}
\label{algo:training}
\end{algorithm}

\subsection{Datasets}
\label{sec:abl:dset}
\subsubsection{Dataset Statistics}\label{apd:dataset}
Table \ref{tab:dset} shows the statistics of the seven datasets used. Note that the Yelp dataset is a heterogeneous graph containing three different views. We choose the edge subset of the ``Review-User-Review (RUR)" view in our experiments.
\begin{table}[H]
    \centering
    \caption{A summary of dataset statistics}
    \label{tab:dset}
    \begin{adjustbox}{width=0.8\columnwidth}
    \begin{tabular}{ccccccc}
    \toprule
                        & \textbf{\# Dim.} & \textbf{\# Nodes} & \textbf{\# Edge} & \textbf{\# Ano. Classes} & \textbf{\# Ano.} & \textbf{PAno. \%} \\ \cmidrule(){1-7}
    \textbf{Photo}      &745       &7,487       &119,043      &2   &369  &4.92    \\
    \textbf{Computers}  &767       &13,381      &245,778      &5   &2,064 &15.42    \\
    \textbf{CS}         &6,805      &18,333      &81,894       &8   &4,159 &22.69   \\  
    \textbf{Yelp}       &32     &45,954      &98,630       &2   &6,677  &14.53  \\
    \textbf{ogbn-arxiv} &128 &169,343 &1,166,243 &4 &27,830 &16.43\\
    \textbf{ogbn-proteins} &8 &132,534 &39,561,252 &2 &10,693 &8.07\\
    \textbf{T-Finance} &10 &39,357 &21,222,543 &2 &1,803 &4.58\\
    \bottomrule    \\         
    \end{tabular}
    \end{adjustbox}
\end{table}

\subsubsection{Dataset Preprocessing}
For the imbalance attributed node classification datasets \textit{Photo}, \textit{Computers}, \textit{CS} and \textit{ogbn-proteins}, we treat each class that has less than 5\% of the total number of nodes of the graph as an anomaly class and the remaining classes as normal. In the ogbn-arxiv dataset, with the original graph containing 40 subclasses, for experimental efficiency, any class representing 3\% to 5\% of the total nodes is considered an anomaly class. Note that the normal class is defined as the combination of all the majority classes, from which labelled normal nodes are randomly and uniformly drawn, and remain consistent through the entire experiment.

For the \textit{Yelp} and \textit{T-Finance} datasets, which only contain binary anomaly labels, we first apply an unsupervised representation learning method Deep Graph Infomax (DGI) for representation learning and then apply k-means clustering to partition the anomaly class into two anomaly subclasses using their node representations. Specifically, our DGI model employs a two-layer GraphSAGE as the representation learner with 128 hidden units per layer. The model is optimised for 1,000 epochs using the Adam optimiser with the learning rate and the batch size set to $1 \times 10^{-3}$ and 512, respectively. To further magnify the distribution shift, instead of employing random sampling, we conduct weighted random sampling to generate training anomaly nodes, where the probability associated with each anomaly node is proportional to its distance from its corresponding cluster centroid.

\subsection{Experimental Protocol} \label{apd:eval}
\begin{algorithm}
\caption{Open-set GAD Experimental Protocol}
\begin{algorithmic}[1]
\REQUIRE $\mathcal{G}, \mathbf{Y}$, and anomaly class index $\mathbf{idx}_{\mathrm{anom}}$.
\ENSURE The overall and unseen AUC-ROC and AUC-PR values.
\STATE $\mathcal{V}_{\mathrm{n}} \leftarrow \{v \mid y_v \notin \mathbf{idx}_{\mathrm{anom}}\}$ 
\STATE $\mathcal{V}^{\mathrm{train}}_{\mathrm{n}} \leftarrow $ Sample the training normal nodes.
\FOR{$c$ in $\mathbf{idx}_{\mathrm{anom}}$}
    \STATE $\mathcal{V}^{\mathrm{seen}}_{\mathrm{a}} \leftarrow \{v \mid y_v == c\}$
    \STATE $\mathcal{V}^{\mathrm{unseen}}_{\mathrm{a}} \leftarrow \{v \mid y_v \text{ not in } \mathbf{idx}_{\mathrm{anom}}\}$
    \STATE $\mathbf{V}_{\mathrm{train}} \leftarrow$ sample from $\mathcal{V}^{\mathrm{seen}}_{\mathrm{a}}$ and $\mathcal{V}^{\mathrm{train}}_{\mathrm{n}}$.
    \STATE $\phi \leftarrow$ Train an anomaly detection network using NSReg as described in Algorithm 1. 
    \STATE $\text{AUC-ROC}_c^{\mathrm{all}}$, $\text{AUC-PR}_c^{\mathrm{all}} \leftarrow$ evaluate $\phi$ using $\mathcal{V} \setminus \mathbf{V}_{\mathrm{train}}$.
    \STATE $\text{AUC-ROC}_c^{\mathrm{useen}}$, $\text{AUC-PR}_c^{\mathrm{unseen}} \leftarrow$ evaluate $\phi$ using $\mathcal{V}_{\mathrm{n}} \cup \mathcal{V}^{\mathrm{unseen}} \setminus \mathbf{V}^{\mathrm{train}}$.
    \STATE Record class AUC-ROC and AUC-PR.
\ENDFOR
\STATE Calculate the average performance ${\text{AUC-ROC}}^{\mathrm{all}}$, ${\text{AUC-PR}}^{\mathrm{all}}$, ${\text{AUC-ROC}}^{\mathrm{unseen}}$, and ${\text{AUC-PR}}^{\mathrm{unseen}}$.
\RETURN ${\text{AUC-ROC}}^{\mathrm{all}}$, ${\text{AUC-PR}}^{\mathrm{all}}, {\text{AUC-ROC}}^{\mathrm{unseen}}$, ${\text{AUC-PR}}^{\mathrm{unseen}}$
\end{algorithmic}
\label{algo:exp}
\end{algorithm}
In this section, we outline the process of conducting open-set GAD experiments and encapsulate it in Algorithm \ref{algo:exp}. For each dataset, we alternate treating each anomaly class as the seen anomaly, with the remaining anomaly classes as unseen, for all anomaly classes. For example, in the CS dataset, 8 out of 15 classes are marked as anomalies. The results for this dataset underwent 5 independent runs of 8 sub-experiments, with each anomaly class treated as `seen' in rotation, totaling 40 sub-experiments for CS.  
Note that the normal class remains consistent across each run, and the labelled normal nodes are uniformly drawn.

\subsection{Implementation Details} 
\label{apd:implem}
\subsubsection{Dependencies}
NSReg is implemented in Python and makes extensive use of Pytorch \citep{pytorch} and Pytorch Geometric \citep{pyg}. We summarise the main scientific computing libraries and their versions used for our implementation as follows:
\begin{itemize}
    \item python==3.8.13
    \item pytorch==1.10.1 (py3.8\_cuda11.3\_cudnn8.2.0\_0)
    \item pytorch\_geometric==2.0.4 (py38\_torch\_1.10.0\_cu113)
    \item numpy==1.23.5
    \item scipy==1.10.0
    \item scikit-learn==1.2.1
    \item cudatoolkit==11.3.1
    \item dgl==1.0.2
\end{itemize}

\subsubsection{Hyperparameter Settings}
For NSReg, in addition to the hyperparameters reported in Sec. \ref{sec:exps}, we set the numbers of neighbours used for GraphSAGE aggregation to 25 and 10 for the first layer and the second layer, respectively, for better runtime efficiency. The same sampling strategy as NSReg is employed for the GNN-adapted competing methods that were designed for Euclidean data. Our anomaly scoring network $\eta$ is a two-layer neural network with 32 hidden units in the hidden layer. BCE, DevNet and PReNet are trained for the same number of epochs (200) as NSReg using the Adam optimiser using the learning rate of $1 \times 10^{-3}$. We keep the default hyperparameter settings for the other baselines except for GraphSMOTE, for which the pretraining and training epoch numbers are both set to 2,500 where the model exhibits convergence. 

For the large-scale datasets ogbn-arxiv and ogbn-proteins, we set the default number of labelled anomalies to 100, and a reduced percentage of labelled normal nodes to 1\% to make it more challenging and practical for real-world applications. For ogbn-proteins, where it is set to 2 to enforce stronger structural regularisation due to its large number of nodes and highly dense connections (i.e., over a hunderd thousand nodes with over 300 average degree). 

\subsubsection{Visualisation Method}
We utilise UMAP \citep{umap}, a widely adopted non-linear invertible dimensionality reduction algorithm, in conjunction with the Matplotlib library to generate contour plots representing anomaly scores. Specifically, we apply UMAP to transform node representations at the anomaly scoring (final) layer of each method into a 2-dimensional representation space. Within the boundaries of the 2D representation space for each method, determined by the maximum values of each dimension, we create a uniform mesh with a step size of 0.1. Subsequently, all points within the mesh are coloured based on the anomaly scores assigned by the final layer, utilising their representations transformed by the inverse transformation of the same UMAP model.

\section{Additional Experimental Results}
\label{sec:apd:res}
\subsection{Detailed Results for NSR as a Plug-and-Play Module}
\label{sec:apd:improve}
We present the complete results of plugging our NSR module into three state-of-the-art supervised AD model, DevNet, DCI and BWGNN, with the original models as the baselines for both all test anomalies and in Table \ref{tab:improve_apd_all} the unseen anomalies in Table \ref{tab:improve_apd_unseen} in terms of both AUC-ROC and AUC-PR. 
Similar substantial performance improvement can be observed as Sec. \ref{sec:main_result}. 

\begin{table}[h]
\centering
\caption{Results of plugging our NSR module into three SOTA supervised AD models DevNet, DCI and BWGNN, with the original models as the baselines on all test anomalies.}
\begin{adjustbox}{width=0.8\columnwidth}
\begin{tabular}{cccccccc}
\toprule     
&Metric & \multicolumn{3}{c}{AUC-ROC} & \multicolumn{3}{c}{AUC-PR} \\  \cmidrule(lr){3-5} \cmidrule(lr){6-8} 
                           &Dataset           & Ori     & +NSR     & Diff.   & Ori     & +NSR     & Diff. \\ \midrule
\multirow{5}{*}{{DevNet}} & Photo     & 0.599   & 0.684   & +0.085$\textcolor{blue}\uparrow$   & 0.223   & 0.424   & +0.201$\textcolor{blue}\uparrow$  \\
&Computers & 0.606   & 0.646   & +0.040$\textcolor{blue}\uparrow$   & 0.284   & 0.340   & +0.056$\textcolor{blue}\uparrow$  \\
& CS        & 0.769   & 0.949   & +0.180$\textcolor{blue}\uparrow$   & 0.684   & 0.872   & +0.188$\textcolor{blue}\uparrow$  \\
& Yelp      & 0.675   & 0.684   & +0.009$\textcolor{blue}\uparrow$   & 0.315   & 0.315   & +0.008$\textcolor{blue}\uparrow$  \\ \cmidrule(lr){2-2} 
 \cmidrule(lr){3-5} \cmidrule(lr){6-8} 
& Avg. Diff.  & \multicolumn{3}{r}{+0.079$\textcolor{blue}\uparrow$}   & \multicolumn{3}{r}{+0.113$\textcolor{blue}\uparrow$} \\ \cmidrule(lr){1-8}
\multirow{5}{*}{{DCI}}  
&  Photo     &0.772	&0.865	&+0.093$\textcolor{blue}\uparrow$	&0.452	&0.577	&+0.125$\textcolor{blue}\uparrow$  \\
& Computers &0.683	&0.738	&+0.055$\textcolor{blue}\uparrow$	&0.427	&0.501	&+0.074$\textcolor{blue}\uparrow$  \\
& CS        &0.856	&0.857	&+0.001$\textcolor{blue}\uparrow$	&0.635	&0.623	&-0.012 \\
& Yelp      &0.689	&0.740	&+0.051$\textcolor{blue}\uparrow$	&0.351	&0.387	&+0.036$\textcolor{blue}\uparrow$ \\ \cmidrule(lr){2-2}  \cmidrule(lr){3-5} \cmidrule(lr){6-8} 
& Avg. Diff.& \multicolumn{3}{r}{+0.050$\textcolor{blue}\uparrow$}   & \multicolumn{3}{r}{+0.056$\textcolor{blue}\uparrow$}  \\ \cmidrule(lr){1-8}
\multirow{5}{*}{{BWGNN}}  & Photo     &0.728	&0.802	&+0.074$\textcolor{blue}\uparrow$	&0.313	&0.530	&+0.217$\textcolor{blue}\uparrow$     \\
& Computers &0.635	&0.726	&+0.091$\textcolor{blue}\uparrow$	&0.348	&0.458	&+0.110$\textcolor{blue}\uparrow$  \\
& CS        &0.845	&0.879	&+0.034$\textcolor{blue}\uparrow$	&0.687	&0.718	&+0.031$\textcolor{blue}\uparrow$  \\
& Yelp      &0.727	&0.747	&+0.020$\textcolor{blue}\uparrow$	&0.366	&0.394	&+0.028$\textcolor{blue}\uparrow$  \\ \cmidrule(lr){2-2}  \cmidrule(lr){3-5} \cmidrule(lr){6-8}
& Avg. Diff.& \multicolumn{3}{r}{+0.055$\textcolor{blue}\uparrow$}   & \multicolumn{3}{r}{+0.097$\textcolor{blue}\uparrow$} \\ \bottomrule
\end{tabular}
\end{adjustbox}
\label{tab:improve_apd_all}
\end{table}

\begin{table}[h]
\centering
\caption{Results of plugging our NSR module into three SOTA supervised AD models DevNet, DCI and BWGNN, with the original models as the baselines on unseen anomalies.}
\begin{adjustbox}{width=0.8\columnwidth}
\begin{tabular}{cccccccc}
\toprule
&Metric & \multicolumn{3}{c}{AUC-ROC} & \multicolumn{3}{c}{AUC-PR} \\  \cmidrule(lr){3-5} \cmidrule(lr){6-8} 
Met.                            &Dataset           & Ori     & +NSR     & Diff.   & Ori     & +NSR     & Diff. \\ \midrule
\multirow{5}{*}{{DevNet}} 
&Photo      &0.468	&0.513	&0.045$\textcolor{blue}\uparrow$	&0.045	&0.092	&0.047$\textcolor{blue}\uparrow$ \\
&Computers  &0.573	&0.623	&0.050$\textcolor{blue}\uparrow$	&0.200	&0.266	&0.066$\textcolor{blue}\uparrow$ \\
&CS         &0.739	&0.944	&0.205$\textcolor{blue}\uparrow$	&0.606	&0.841	&0.235$\textcolor{blue}\uparrow$ \\
&Yelp       &0.621	&0.632	&0.011$\textcolor{blue}\uparrow$	&0.142	&0.146	&0.004$\textcolor{blue}\uparrow$\\ \cmidrule(lr){2-2} 
 \cmidrule(lr){3-5} \cmidrule(lr){6-8} 
& Avg. Diff.  & \multicolumn{3}{r}{0.078$\textcolor{blue}\uparrow$}   & \multicolumn{3}{r}{0.088$\textcolor{blue}\uparrow$} \\ \cmidrule(lr){1-8}
\multirow{5}{*}{{DCI}}  
&Photo     &0.614	&0.780	&0.166$\textcolor{blue}\uparrow$	&0.083	&0.201	&0.118$\textcolor{blue}\uparrow$      \\
&Computers &0.629	&0.693	&0.064$\textcolor{blue}\uparrow$	&0.288	&0.380	&0.092$\textcolor{blue}\uparrow$ \\
&CS        &0.847	&0.847	&0.000	&0.558	&0.547	&-0.011 \\
& Yelp     &0.637	&0.699	&0.062$\textcolor{blue}\uparrow$	&0.157	&0.190	&0.033$\textcolor{blue}\uparrow$  \\ \cmidrule(lr){2-2}  \cmidrule(lr){3-5} \cmidrule(lr){6-8} 
& Avg. Diff.& \multicolumn{3}{r}{0.073$\textcolor{blue}\uparrow$}   & \multicolumn{3}{r}{0.058$\textcolor{blue}\uparrow$}  \\ \cmidrule(lr){1-8}
\multirow{5}{*}{{BWGNN}}  
&Photo     &0.598	&0.628	&0.030$\textcolor{blue}\uparrow$	&0.068	&0.078	&0.010$\textcolor{blue}\uparrow$     \\
&Computers &0.570	&0.662	&0.092$\textcolor{blue}\uparrow$	&0.209	&0.272	&0.063$\textcolor{blue}\uparrow$ \\
&CS        &0.829	&0.873	&0.044$\textcolor{blue}\uparrow$	&0.620	&0.678	&0.058$\textcolor{blue}\uparrow$ \\
& Yelp     &0.674	&0.703	&0.029$\textcolor{blue}\uparrow$	&0.167	&0.188	&0.021$\textcolor{blue}\uparrow$ \\ \cmidrule(lr){2-2}  \cmidrule(lr){3-5} \cmidrule(lr){6-8}
& Avg. Diff.& \multicolumn{3}{r}{0.049$\textcolor{blue}\uparrow$}   & \multicolumn{3}{r}{0.038$\textcolor{blue}\uparrow$} \\ \bottomrule
\end{tabular}
\end{adjustbox}
\label{tab:improve_apd_unseen}
\end{table}

\vspace{10cm}
\subsection{Detailed Ablation Study Results.} 
\label{apd:ablation}
We present the comprehensive results of the ablation study in Table \ref{tab:abl_apd_both} and \ref{tab:abl_apd_auroc}, corresponding to Table \ref{tab:abl} in the paper. The performance in AUC-ROC shows a similar trend as AUC-PR, aligning with our observations and findings in Sec. \ref{sec:abl}.

\begin{table}[h]
\centering
\caption{Our proposed NSR module vs other regularisation methods for all test anomalies.}
\begin{adjustbox}{width=0.9\columnwidth}
\begin{tabular}{ccccccc}
\toprule
& Dataset  &SAD  &CL (unsup) &CL (s) &NSR (dual) & NSR  \\ \midrule
\multirow{5}{*}{AUC-ROC}  
& Photo         &0.599±0.134    &0.807±0.010    &0.891±0.010    &0.894±0.016	&\bf{0.908±0.016} \\ 
& Computers &0.644±0.126    &0.743±0.012    &0.781±0.022    &0.778±0.021	&\bf{0.797±0.015}  \\
& CS        &0.819±0.014    &0.894±0.006    &0.927±0.012	&0.927±0.022    &\bf{0.957±0.007} \\
& Yelp      &0.522±0.041	&0.731±0.014    &0.665±0.072    &0.732±0.016	&\bf{0.734±0.012} \\ \cmidrule(lr){2-7}
& average   &0.646±0.079	&0.794±0.011	&0.816±0.029	&0.833±0.019	&\bf{0.849±0.013} \\ \midrule
\multirow{5}{*}{AUC-RR}   
& Photo     &0.269±0.215    &0.513±0.018    &0.583±0.012    &0.623±0.029     &\bf{0.640±0.036} \\
& Computers &0.381±0.190    &0.492±0.014    &0.539±0.024	&0.539±0.023     &\bf{0.559±0.018} \\
& CS        &0.700±0.016    &0.795±0.009	&0.843±0.020	&0.858±0.042     &\bf{0.889±0.016} \\
& Yelp      &0.175±0.046    &0.392±0.013	&0.330±0.078    &0.399±0.021     &\bf{0.398±0.014} \\ \cmidrule(lr){2-7}
& Average   &0.646±0.079	&0.794±0.011	&0.816±0.029	&0.833±0.019	 &\bf{0.849±0.013}\\ \bottomrule
\end{tabular}
\end{adjustbox}
\label{tab:abl_apd_both}
\end{table}
\begin{table}[h]
\centering
\caption{Our proposed NSR module vs other regularisation methods for unseen anomalies.}
\begin{adjustbox}{width=0.9\columnwidth}
\begin{tabular}{ccccccc}
\toprule
& Dataset  &SAD  &CL (unsup) &CL (s) &NSR (dual) & NSR  \\ \midrule
\multirow{5}{*}{AUC-ROC}  
& Photo     &0.516±0.047	&0.652±0.018	&0.807±0.017	&0.810±0.031	&\bf{0.836±0.031}\\ 
& Computers &0.614±0.103	&0.685±0.015	&0.733±0.027	&0.730±0.025	&\bf{0.752±0.019}\\
& CS         &0.797±0.016	&0.883±0.006	&0.920±0.013	&0.924±0.021	&\bf{0.953±0.008}\\
& Yelp      &0.511±0.026	&0.677±0.023	&0.613±0.071	&0.679±0.018	&\bf{0.690±0.025}\\ \cmidrule(lr){2-7}
& average    &0.610±0.048	&0.724±0.016	&0.768±0.032	&0.786±0.024	&\bf{0.808±0.021}\\ \midrule
\multirow{5}{*}{AUC-RR}   
& Photo     &0.053±0.004	&0.068±0.004	&0.133±0.019	&0.207±0.049	&\bf{0.221±0.057}\\
& Computers &0.285±0.134	&0.318±0.018	&0.379±0.032	&0.389±0.028	&\bf{0.417±0.024}\\
& CS        &0.631±0.019	&0.743±0.010	&0.797±0.025	&0.834±0.046 	&\bf{0.866±0.021}\\
& Yelp      &0.090±0.017	&0.178±0.011	&0.152±0.045	&0.190±0.023 	&\bf{0.196±0.020} \\ \cmidrule(lr){2-7}
& Average   &0.265±0.044	&0.327±0.011	&0.365±0.030	&0.405±0.037	&\bf{0.425±0.031} \\ \bottomrule
\end{tabular}
\end{adjustbox}
\label{tab:abl_apd_auroc}
\end{table}

\subsection{Detailed Large-scale GAD Results}
\label{sec:apd:large}
In this section, we present the complete results for large-scale GAD in Tables \ref{tab:large-scale-apd-all} and \ref{tab:large-scale-apd-unseen}. We observe NSReg consistently outperforms the baseline methods by a large margin on unseen anomalies, similar to our discussion in Sec. \ref{sec:large-scale}.

\begin{table}[h]
\caption{Results on large-scale graph datasets for all test anomalies, where ``-'' denotes unavailable results due to out of memory.}
\centering
\vspace{5pt}
\begin{adjustbox}{width=0.7\columnwidth}
\begin{tabular}{cccccc}
\toprule 
Metric &Datasets & ogbn-arxiv & ogbn-proteins &T-Finance & avg. \\ \midrule
\multirow{10}{*}{\begin{tabular}[c]{@{}c@{}c@{}}AUC-\\ ROC\\(All) \end{tabular}} 
&PReNet   &0.581±0.006	&0.613±0.035	&0.892±0.017	&0.695±0.019  \\
&DevNet   &0.601±0.015	&0.622±0.057	&0.654±0.210	&0.626±0.094  \\
&DCI      &0.566±0.041	&0.815±0.037	&0.763± 0.111	&0.715±0.063 \\
&BWGNN    &0.587±0.013	&0.727±0.067	&0.922±0.011	&0.745±0.030  \\
&AMNet    &0.600±0.049	&0.711±0.077	&0.889±0.024	&0.733±0.050  \\
&GHRN     &0.588±0.009	&0.674±0.019	&0.923±0.010	&0.728±0.013\\
&G.ENS    &-            &-          &0.847±0.087  &-          \\
&G.SMOTE  &-            &-          &0.875±0.016 &-          \\
&BCE      &0.592±0.004	&0.568±0.020	&0.922±0.011	&0.694±0.012    \\ \cmidrule(lr){2-6}  
&NSReg    &\bf{0.659±0.012}	&\bf{0.843±0.029}	&\bf{0.929±0.007} &\bf{0.810±0.016}  \\  \cmidrule{1-6}
\multirow{10}{*}{{\begin{tabular}[c]{@{}c@{}c@{}}AUC-\\ PR\\(All) \end{tabular}}} 
&PReNet  &0.288±0.004	&0.432±0.028	&0.571±0.140	&0.430±0.057   \\
&DevNet  &0.304±0.009	&0.436±0.017	&0.323±0.327	&0.354±0.118    \\
&DCI     &0.275±0.033	&0.597±0.062	&0.264±0.240	&0.379±0.112\\
&BWGNN   &0.263±0.009	&0.582±0.026	&0.746±0.023	&0.530±0.019\\
&AMNet   &0.272±0.053	&0.576±0.053	&0.644±0.046	&0.497±0.051\\
&GHRN    &0.271±0.007	&0.564±0.005	&0.727±0.031	&0.521±0.014\\
&G.ENS   &-           &-            &0.332±0.076    &-            \\
&G.SMOTE &-           &-            &0.573±0.077    &-            \\ 
&BCE     &0.310±0.003	&0.524±0.008	&0.726±0.034	&0.520±0.015   \\  \cmidrule(lr){2-6}
&NSReg   &\bf{0.336±0.006}	&\bf{0.723±0.020} &\bf{0.757±0.020} &\bf{0.605±0.125} \\
\bottomrule
\end{tabular}
\end{adjustbox}
\label{tab:large-scale-apd-all}
\end{table}

\begin{table}[h]
\caption{Results on large-scale graph datasets for unseen anomalies, where ``-'' denotes unavailable results due to out of memory.}
\centering
\vspace{5pt}
\begin{adjustbox}{width=0.7\columnwidth}
\begin{tabular}{cccccc}
\toprule  
Metric &Datasets & ogbn-arxiv & ogbn-proteins &T-Finance & avg. \\ \midrule
\multirow{10}{*}{\begin{tabular}[c]{@{}c@{}c@{}}AUC-\\ ROC\\(Unseen)\end{tabular}} & PReNet   &0.465±0.008	&00.250±0.058	&00.881±0.029	&0.532±0.032\\
&DevNet  &0.494±0.019	&0.264±0.112	&0.642±0.227	&0.467±0.119\\
&DCI     &0.465±0.045	&0.672±0.095	&0.742±0.106	&0.626±0.082 \\
&BWGNN   &0.499±0.014	&0.429±0.156	&0.903±0.023	&0.610±0.064    \\
&AMNet   &0.519±0.053	&0.339±0.263	&0.872±0.045	&0.577±0.120   \\
&GHRN    &0.493±0.012	&0.309±0.043	&0.908±0.012	&0.570±0.022\\
&G.ENS   &-           &- &0.835±0.100 &-      \\
&G.SMOTE &-           &- &0.878±0.016 &-      \\
&BCE     &0.478±0.005	&0.120±0.033	&0.921±0.010	&0.506±0.016 \\ \cmidrule(lr){2-6}
&NSReg   &\bf{0.570±0.016}	&\bf{0.748±0.047}	&\bf{0.928±0.006}	&\bf{0.749±0.023}\\ \midrule
\multirow{10}{*}{\begin{tabular}[c]{@{}c@{}c@{}}AUC-\\ PR\\(Unseen) \end{tabular}} 
&PReNet   &0.124±0.002	&0.029±0.004	&0.431±0.147	&0.195±0.051     \\
&DevNet   &0.142±0.006	&0.032±0.007	&0.284±0.306	&0.153±0.106      \\
&DCI      &0.125±0.011	&0.182±0.172	&0.185±0.221	&0.164±0.135\\
&BWGNN    &0.135±0.004	&0.058±0.018	&0.609±0.029	&0.267±0.017        \\
&AMNet    &0.141±0.024	&0.048±0.030	&0.461±0.067	&0.217±0.040     \\
&GHRN     &0.132±0.003	&0.044±0.002	&0.583±0.057	&0.238±0.021\\
&G.ENS    &-          &-            &0.201±0.061    &-    \\
&G.SMOTE  &-          &-            &0.459±0.095    &-    \\ 
&BCE      &0.136±0.002	&0.025±0.001	&0.629±0.047    &0.263±0.017    \\ \cmidrule(lr){2-6}
&NSReg    &\bf{0.165±0.005}	&\bf{0.488±0.042}	&\bf{0.669±0.024}	&\bf{0.441±0.024}   \\ 
\bottomrule
\end{tabular}
\end{adjustbox}
\label{tab:large-scale-apd-unseen}
\end{table}

\begin{table}[h]
\centering
\caption{NSReg vs. NSReg considering connected normal-anomaly relation (NSReg + NA).}
\begin{adjustbox}{width=0.7\linewidth}
\begin{tabular}{cccccc}
\toprule
                         &           & \multicolumn{2}{c}{AUC-ROC} & \multicolumn{2}{c}{AUC-PR}                        \\ \cmidrule(lr){3-4} \cmidrule(lr){5-6}
                         &           & NSReg + NA         & NSReg         &  NSReg + NA                                & NSReg         \\ \midrule
                         & Photo     & 0.860±0.017  & 0.908±0.016  & 0.561±0.019                         & 0.640±0.036 \\
                         & computers & 0.761±0.013  & 0.797±0.015  & 0.538±0.018                         & 0.559±0.018 \\
                         & CS   & 0.965±0.005  & 0.957±0.007  & 0.904±0.014                         & 0.889±0.016 \\ 
                         & Yelp      & 0.738±0.004  & 0.734±0.012  & 0.401±0.009 & 0.398±0.014 \\ \cmidrule(lr){2-6}
\multirow{-5}{*}{all}    & Average   & 0.831±0.010  & 0.849±0.013  & 0.601±0.015                         & 0.622±0.021 \\ \cmidrule(lr){1-6}
                         & Photo     & 0.747±0.032  & 0.836±0.031  & 0.107±0.018                         & 0.221±0.057 \\
                         & Computers & 0.708±0.015  & 0.752±0.019  & 0.392±0.025                         & 0.417±0.024 \\
                         & CS  & 0.962±0.006  & 0.953±0.008  &0.884±0.018 & 0.866±0.021 \\
                         & Yelp      & 0.689±0.009  & 0.690±0.025  & 0.193±0.010 & 0.196±0.020 \\ \cmidrule(lr){2-6}
\multirow{-5}{*}{unseen} & Average   & 0.777±0.016  & 0.808±0.021  & 0.394±0.018                         & 0.425±0.031 \\ 
\bottomrule
\end{tabular}
\end{adjustbox}
\label{tab:na}
\end{table}

\vspace{10cm}

\subsection{Effect of Including Normal-anomaly Relations}
\label{sec:apd:na}
This section investigates the effect of including an additional type of normal relation that connects normal and anomaly nodes in the relation modelling of NSReg. We report the GAD performance for all test anomalies and the unseen anomalies in both AUC-ROC and AUC-PR in Table \ref{tab:na}. We found that the inclusion of connected normal-anomaly relations is less effective than the default NSReg in terms of average performance, although it can achieve comparable performance on the CS and Yelp datasets. This is because the structural distinction between the normal and the seen anomalies is usually sufficiently preserved by the GAD loss. Additionally, it could introduce unnecessary complexity into the hierarchy of the relation modelling, resulting in less effective modelling of the normality that is not addressed by the GAD loss.

\subsection{Detailed Hyperparameter Analysis}\label{app:hyperparameter}
\begin{figure}[h]
    \centering
    \includegraphics[width=0.5\columnwidth]{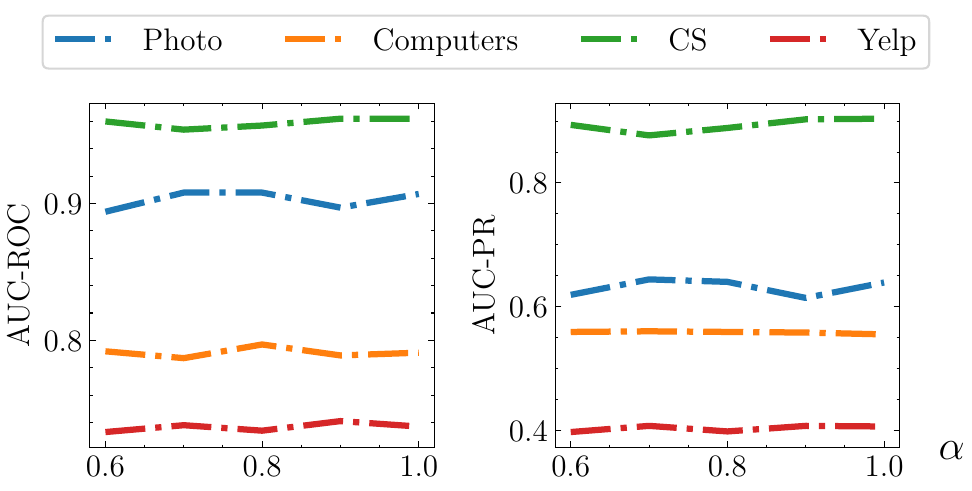}
    \caption{Overall GAD performance of NSReg w.r.t $\alpha$.}
    \label{fig:apd_ps_alpha}
\end{figure}

\begin{figure}[h]
    \centering
    \includegraphics[width=0.5\columnwidth]{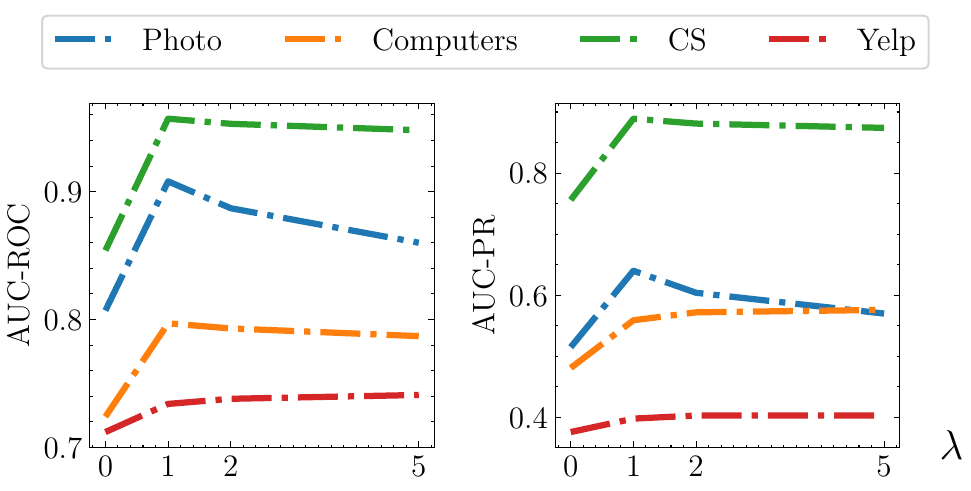}
    \caption{Overall GAD performance of NSReg w.r.t $\lambda$.}
    \label{fig:apd_ps_lambda}
\end{figure}

We report the GAD performance on all test anomalies in both AUC-ROC and AUC-PR with respect to $\lambda$ and $\alpha$ in Figure \ref{fig:apd_ps_alpha} and Figure \ref{fig:apd_ps_lambda}, respectively. The results demonstrate stability across a wide spectrum, indicating the robustness of NSReg with respect to their settings. 


\subsection{Detailed Data Efficiency Results}\label{apd:de}
\begin{figure*}[h]
    \centering
    \includegraphics[width=1.0\linewidth]{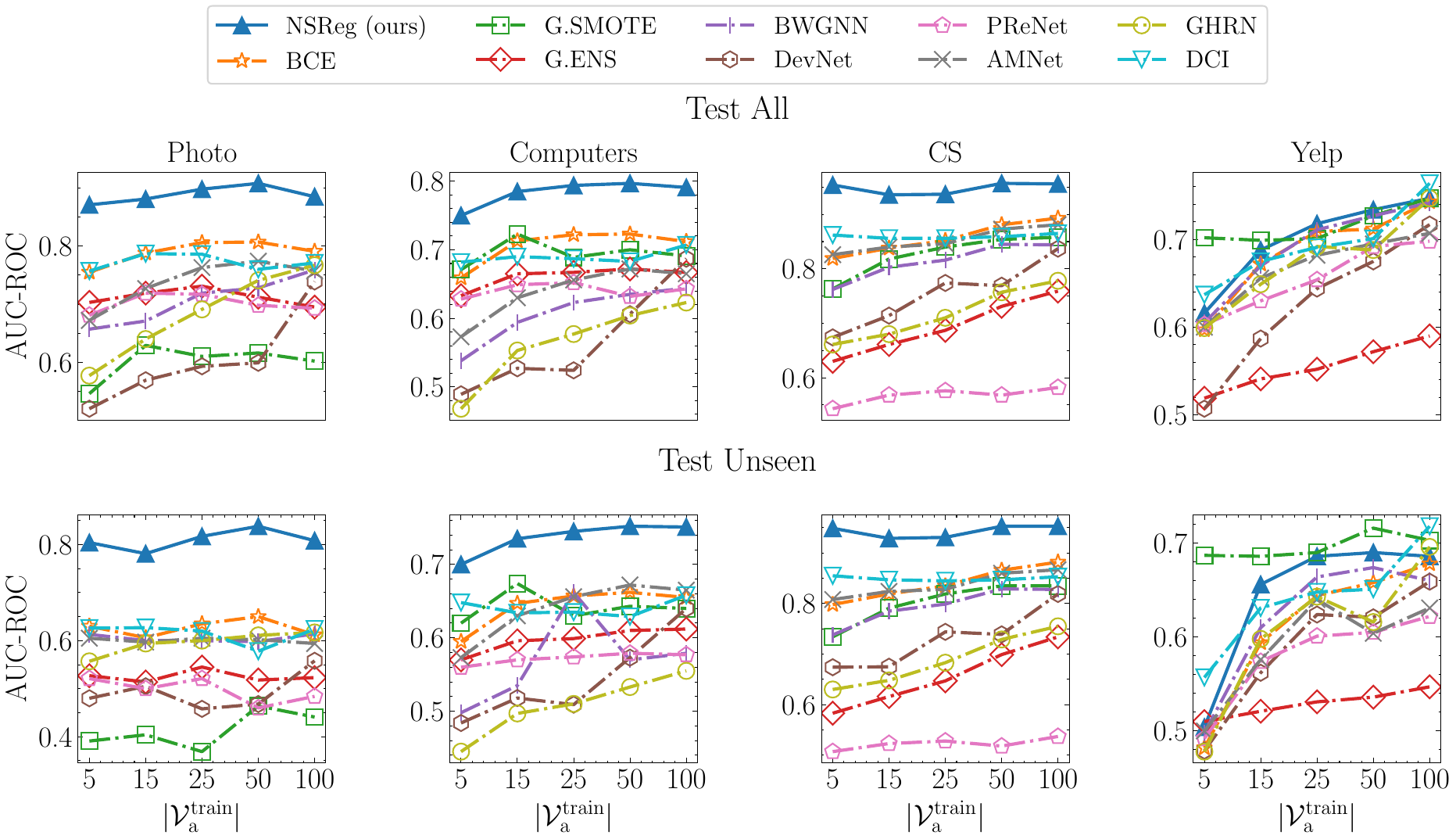}
    \caption{Data efficiency of NSReg and the competing methods in utilising labelled data in AUC-ROC for both all test anomalies and the unseen anomalies.}
    \label{fig:de_apd_aupr}
\end{figure*}

\begin{figure*}[t]
    \centering
    \includegraphics[width=1.0\linewidth]{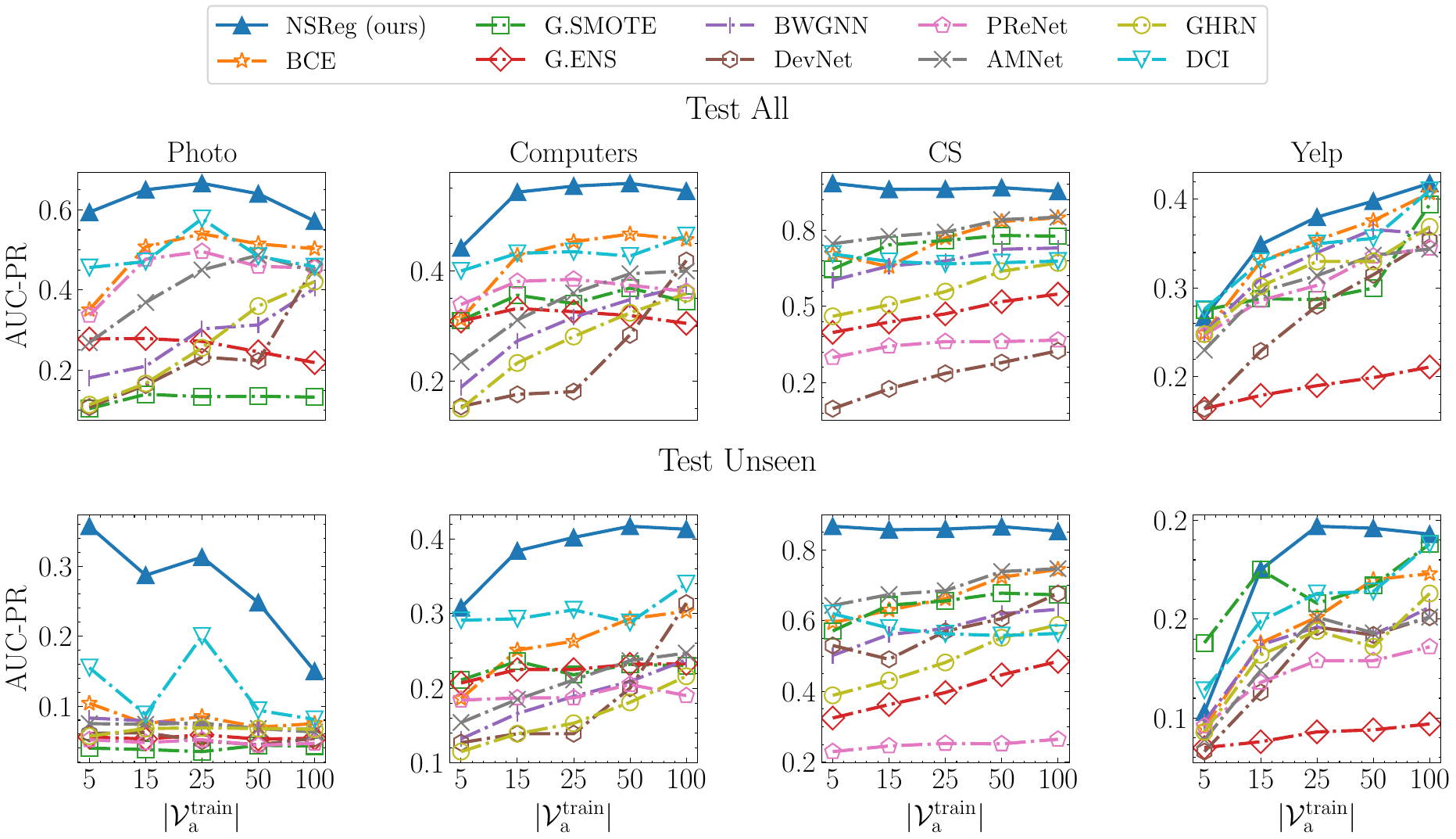}
    \caption{Data efficiency of NSReg and the competing methods in utilising labelled data in AUC-PR for both all test anomalies and the unseen anomalies.}
    \label{fig:de_apd_auroc}
\end{figure*}

In this section, our primary focus is on showing the data efficiency performance of NSReg in terms of AUC-ROC, which complements the discussion of AUC-PR in Sec. \ref{sec:main_result}. As illustrated in Figure \ref{fig:de_apd_aupr} and Figure \ref{fig:de_apd_auroc}, NSReg consistently achieves remarkable performance similar to AUC-PR. Specifically, across the entire spectrum, NSReg outperforms the baselines on both overall and unseen anomaly detection for the Photo, Computers, and CS datasets. For the Yelp dataset, NSReg either outperforms or achieves comparable performance, particularly when the number of labelled training anomalies exceeds 5. While GraphSMOTE slightly outperforms NSReg in unseen anomaly detection, its effectiveness in the seen anomalies is limited, resulting in less effective overall detection performance.

\end{document}